\documentclass[10pt,twocolumn,letterpaper]{article}
\makeatletter
\providecommand{\@LN}[2]{}
\makeatother

\usepackage[pagenumbers]{cvpr} % To force page numbers, e.g. for an arXiv version

\definecolor{cvprblue}{rgb}{0.21,0.49,0.74}
\usepackage[pagebackref,breaklinks,colorlinks,allcolors=cvprblue]{hyperref}
\usepackage{fixltx2e}
\usepackage{amssymb}
\usepackage{multicol}
\usepackage{wrapfig}
\usepackage{graphicx}
\usepackage[linesnumbered, boxed, ruled]{algorithm2e}
\usepackage{cuted}
\usepackage{bm}
\usepackage{threeparttable}
\usepackage{dblfloatfix} % Fixes the placement of two-column floats

\newcommand{\FDD}{FD\textsubscript{DINOv2}\hspace{1.5pt}}

\def\paperID{5801} % *** Enter the Paper ID here
\def\confName{CVPR}
\def\confYear{2025}

\title{Gradient-Free Classifier Guidance for Diffusion Model Sampling}

\author{
Rahul Shenoy$^*$\textsuperscript{1}, Zhihong Pan$^*$\textsuperscript{1}, Kaushik Balakrishnan\textsuperscript{1}\\
Qisen Cheng\textsuperscript{1}, Yongmoon Jeon\textsuperscript{2}, Heejune Yang\textsuperscript{2}, Jaewon Kim\textsuperscript{2} \vspace{3pt}\\
\textsuperscript{1}Samsung Display America Lab, San Jose, CA, USA\\
\textsuperscript{2}Samsung Display Co., Yongin-si, Gyeonggi-do, South Korea
}

\begin{document}
\maketitle
\def\thefootnote{*}\footnotetext{These authors contributed equally to this work}\def\thefootnote{\arabic{footnote}}

\begin{strip}
  \centering
  \vspace{-30pt}
  \includegraphics{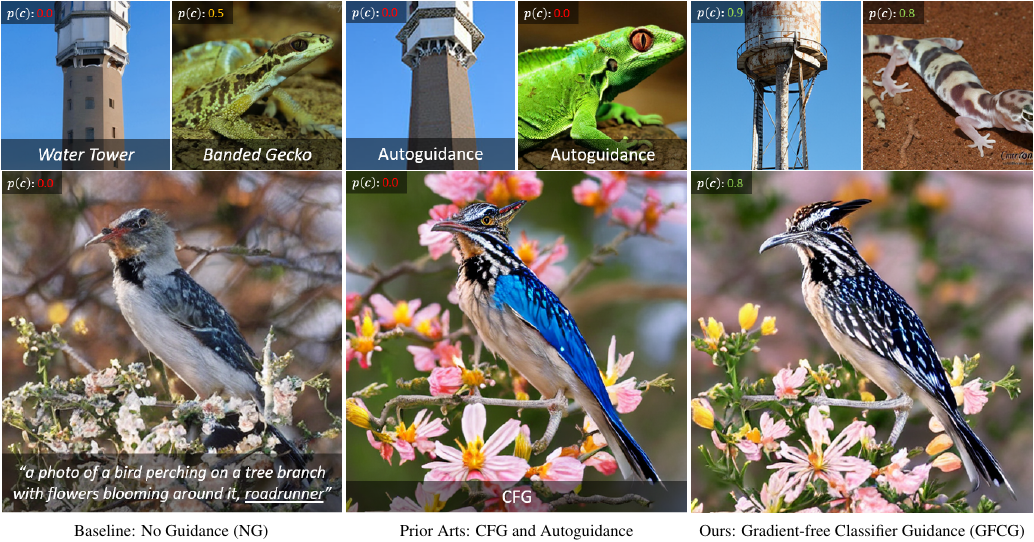} 
  \vspace{-15pt}
  \captionof{figure}{Qualitative comparisons of the proposed gradient-free classifier guidance (GFCG) with other methods, including no-guidance baseline, classifier-free guidance (CFG)~\cite{ho_nipsw_2021} (SD 1.5~\cite{sd_github_2022}) and Autoguidance~\cite{karras_arxiv_2024} (EDM2~\cite{karras_cvpr_2024}).  Our GFCG uses the power of a pre-trained classifier to improve the classification accuracy of sampled images significantly without the burden of gradient descent and loss of diversity.}
  \vspace{-5pt}
  \label{fig:front}
\end{strip}

\begin{abstract}
Image generation using diffusion models have demonstrated outstanding learning capabilities, effectively capturing the full distribution of the training dataset. They are known to generate wide variations in sampled images, albeit with a trade-off in image fidelity.  Guided sampling methods, such as classifier guidance (CG) and classifier-free guidance (CFG), focus sampling in well-learned high-probability regions to generate images of high fidelity, but each has its limitations.  CG is computationally expensive due to the use of back-propagation for classifier gradient descent, while CFG, being gradient-free, is more efficient but compromises class label alignment compared to CG.  In this work, we propose an efficient guidance method that fully utilizes a pre-trained classifier without using gradient descent.  By using the classifier solely in inference mode, a time-adaptive reference class label and corresponding guidance scale are determined at each time step for guided sampling.  Experiments on both class-conditioned and text-to-image generation diffusion models demonstrate that the proposed Gradient-free Classifier Guidance (GFCG) method consistently improves class prediction accuracy. We also show GFCG to be complementary to other guided sampling methods like CFG. When combined with the state-of-the-art Autoguidance (ATG), without additional computational overhead, it enhances image fidelity while preserving diversity.  For ImageNet 512$\times$512, we achieve a record \FDD of 23.09, while simultaneously attaining a higher classification Precision (94.3\%) compared to ATG (90.2\%).
\end{abstract}
\section{Introduction}
\label{sec:intro}

Denoising diffusion models~\cite{ho_nips_2020, song_nips_2019, song_iclr_2020, song_iclr_2021}, the latest popular generative models, have demonstrated exceptional performance across various domains\cite{kong_iclr_2021, luo_cvpr_2021, ho_iclrw_2022, ren_icassp_2023, xu_icml_2023}, particularly in image generation~\cite{nichol_arxiv_2022, rombach_cvpr_2022, saharia_nips_2022}. By leveraging a learned denoising network, these models iteratively refine outputs to produce diverse, high-quality images.  To enhance control over image generation,  the denoiser is often trained with specific conditions to generate images with desired properties, commonly employing class labels~\cite{ho_nipsw_2021} or text prompt embeddings~\cite{saharia_nips_2022, nichol_arxiv_2022}, as well as other types of image conditions~\cite{saharia_nipsw_2021}.

Guiding an unconditional model to generate images of a specific class can be accomplished using classifier guidance (CG)~\cite{dhariwal_nips_2021}. During the iterative image generation process, CG steers the model towards outputs that align with a designated class by incorporating the classifier's gradient at each step.  This approach not only improves image fidelity but also applies to other attributes beyond class~\cite{pan_wacv_2023, yu_iccv_2023, bansal_cvpr_2023}.  Despite its effectiveness in generating images with desired attributes, gradient-based guidance methods such as CG are computationally inefficient due to the time-consuming back-propagation required at each sampling step, often multiple times per step~\cite{ye_arxiv_2024}. Although CG enhances image quality, indicated by increased precision scores when tested on synthetic images using a real classifier~\cite{dhariwal_nips_2021}, it also compromises image diversity, as evidenced by lower FID scores~\cite{heusel_nips_2017} and decreased recall when evaluated on real images using a classifier trained on synthetic images.

Classifier-free guidance (CFG)~\cite{ho_nipsw_2021} is the first gradient-free guidance method that improves image quality without requiring a classifier.  This is achieved by generating an unconditional sample  and using it as a bad sample to avoid.  CFG experiences a similar trade-off between sample quality and diversity as CG. Autoguidance (ATG)~\cite{karras_arxiv_2024}, a recent gradient-free guidance method, proposes using a sample from a bad version of the model instead of the bad unconditional sample used in CFG.  It guides sampling to high-probability regions of the data distribution without reducing diversity, achieving state-of-the-art performances in both FID~\cite{heusel_nips_2017} and FD\textsubscript{DINOv2}~\cite{stein_nips_2024} metrics.  However, the trade-off in image quality as evaluated using classification metrics, is not addressed.

In this work, we propose a new gradient-free guidance method that uses the sample from a reference class ($c_{ref}$) as the undesired sample to avoid.  Unlike CFG, where an empty class $\emptyset$ or a separate unconditional model is fixed throughout the sampling process, a pre-trained classifier is used to adaptively determine the reference class and guidance scale $\omega$ based on classier predictions.  Compared to gradient-based classifier guidance, the proposed gradient-free classifier guidance (GFCG) offers similar benefit of better alignment between generated images and class label conditions, without the computational expense of classifier gradient descent.  Additionally, it is complementary to other guidance methods and can be combined with them to improve image quality without trade-off in diversity. We investigate two combination methods for their efficiency and effectiveness. For the mixed guidance, it combines the guidance from two methods temporally resulting in no additional overhead in the number of function evaluations (NFEs). Alternatively, the additive guidance method combines the guidance terms from two methods, achieving the best performance in both image quality and diversity at the cost of more computations, as the guidance terms are calculated separately. Extensive experiments are conducted on class-conditional and text-to-image diffusion models to demonstrate the benefits of GFCG in improving image fidelity with minimal trade-off in diversity.  Extensive ablation studies are conducted to provide insights into optimal settings for GFCG alone and in combination with other methods like CFG and ATG. As shown in Figure ~\ref{fig:front}, our results significantly improve classification accuracy for different models.  For the bottom text-to-image generation, GFCG successfully generates the correct bird species while maintaining coherence with the detailed description.

In summary, the proposed GFCG method has the following key advantages:
\begin{itemize}

\setlength\itemsep{0.05em}
\item[$\bullet$] It is the first known work of using a classifier for gradient-free guidance in diffusion sampling, leveraging a pre-trained classifier to adaptively determine both the reference class and guidance scale during sampling.

\item[$\bullet$] It significantly enhances image fidelity for both class-conditional and text-to-image models.

\item[$\bullet$] It is complementary to other gradient-free guidance methods. When combined with ATG, it establishes a new state-of-the-art performance in metrics measuring both sample image fidelity and diversity.

\end{itemize}

\section{Related Works}
\label{sec:rwork}

\noindent\textbf{Gradient-based Diffusion Guidance} Dhariwal \etal~\cite{dhariwal_nips_2021} were the first to propose gradient-based guided sampling for diffusion model.  In addition to the trade-off in image diversity and time-consuming classifier gradient calculation, it requires additional training of a noise-conditional classifier.  For works following that, there are two main focuses, to generalize the guidance from classifier loss to any differentiable loss functions~\cite{bansal_cvpr_2023}, and apply it through an existing differentiable target predictor without involving any additional training using noisy images~\cite{chung_arxiv_2022}. Additionally, various techniques have been proposed to improve guidance effects, including guidance with multiple samples from Monte Carlo (MC) method~\cite{song_icml_2023}, repetitive guided sampling at each time step~\cite{yu_iccv_2023}, and computing the loss function gradient to the estimated denoised image~\cite{he_arxiv_2023}.  Ye \etal~\cite{ye_arxiv_2024} introduced a unified framework which encompass the prior works as special cases.  By optimizing multiple hyperparameters in the unified framework, it is able to achieve improved performance over prior works in both FID and classification accuracy metrics.  While the proposed techniques like MC sampling and recurrent guidance are shown beneficial to guidance effects, they all come with additional computational steps to different extents, making it even worse with already expensive gradient calculations.

\noindent\textbf{Gradient-free Diffusion Guidance} Ho \etal~\cite{ho_nipsw_2021} introduced the first gradient-free guidance method, classifier-free guidance (CFG), to improve image quality in diffusion sampling without the need of a classifier.  At each sampling step, it modifies the class conditional sample by enhancing the contrast between it and the corresponding unconditional sample.  CFG enhances sample quality by guiding samples toward high-probability regions but suffers reduction in diversity for over-correction in class label alignment.  Karras \etal~\cite{karras_arxiv_2024} proposed a new method, Autoguidance (ATG), to improve sample quality with fewer loss in diversity.  Using a smaller model trained from the same dataset with less learning iterations, it generates a bad reference sample for the main sample to steering away from.  However, the samples are not assessed using classification accuracy so it is not clear if there is trade-off in image quality in terms of alignment with class labels.  This is investigated in our work as part of a comprehensive set of experiments.  There is another line of gradient-free methods which work on generating a bad reference sample by exploiting the self-attention layers of the diffusion network, including blurring pixels with high self-attention~\cite{hong_iccv_2023}, adding Gaussian blur to weights of an intermediate self-attention layer~\cite{hong_arxiv_2024}, and replacing an intermediate self-attention map with an identity matrix~\cite{ahn_arxiv_2024}.  There is no known comparisons between these self-attention based methods and the latest ATG. In addition to the innovative gradient-free classifier method, to our best knowledge, we are the first to conduct comparison studies for the full range of gradient-free guidance methods.

\section{Methods}
\subsection{Diffusion Models}

Diffusion models are a class of generative models that generate data following an iterative denoising process \cite{ho_nips_2020, song_iclr_2021, song_iclr_2020}.
This involves a forward process where noise is ingested to data over a sequence of time steps to render them indistinguishable from
Gaussian noise, and a backward denoising process where the noise is removed following the reverse sequence until noise-free data is
recovered. The forward process is governed by the stochastic differential equation (SDE):
\begin{equation} 
d{\bf x} = {\bf f}({\bf x}, t)dt + g(t) d{\bf w},
\end{equation}
where ${\bf x}$ is the data, $t \in [0, T]$ is the time step, and ${\bf f}$ and $g$ are predefined functions that govern the noise schedule, and $d{\bf w}$ is a standard
Wiener process. The denoising process is governed by the reverse SDE:
\begin{equation}  
d{\bf x} = \left[{\bf f}({\bf x}, t) - g(t)^2 \nabla_{{\bf x}} \text{log} p_t ({\bf x}) \right] dt + g(t)d\overline{{\bf w}}, 
\end{equation}
where $\nabla_{{\bf x}} \text{log} p_t({\bf x})$ is the score function and $d\overline{{\bf w}}$ is the standard Wiener process for the reverse steps. 

In diffusion models, the score function is parameterized by a deep neural network with parameters $\theta$, and represented as $D_{\theta}({\bf x}, t) \approx \nabla_{{\bf x}} \text{log} p_t({\bf x})$.
Conditioning variables such as class label or text prompt, denoted as $c$, can also be included and in this setting $D_{\theta}({\bf x}, t, c) \approx \nabla_{{\bf x}} \text{log} p_t({\bf x}|c)$. 
During the reverse denoising process, to improve the quality of data generation, classifier-free guidance (CFG) is widely used \cite{ho_nipsw_2021}:
\begin{eqnarray}\label{eq:g}
d{\bf x} = \left[{\bf f}({\bf x}, t) - g(t)^2 \widehat{D}({\bf x}, t, c) \right]dt + g(t)d\overline{{\bf w}}, \nonumber \\
\widehat{D}({\bf x}, t, c) = \omega D^{m}_{\theta}({\bf x}, t, c) - (\omega-1) D^{g}_{\phi}({\bf x}, t), 
\end{eqnarray}
where $D^{m}$ and $D^{g}$ are the main and guidance networks, parameterized by neural network weights $\theta$ and $\phi$, respectively.
Here $D^{g}$ could be an unconditional diffusion model as in some implementations.  In others, both $D^{m}$ and $D^{g}$ may be conditional neural networks where a null ($\emptyset$) class is used as the reference for CFG, replacing $D^{g}_{\phi}({\bf x}, t)$ in Equation \ref{eq:g} 
with $D^{g}_{\phi}({\bf x}, t, \emptyset)$.  Furthermore, $\omega$ is a hyperparameter referred to as the ``guidance scale" where a scale of 1 means no guidance.   Guided sampling improves the quality of data generation, albeit trading-off diversity \cite{ho_nipsw_2021}.
%Other recent works have also considered diffusion guidance by applying Gaussian blur to attention maps \cite{hong_arxiv_2024}.

\begin{figure*}
  \centering
  \includegraphics[width=0.9\linewidth]{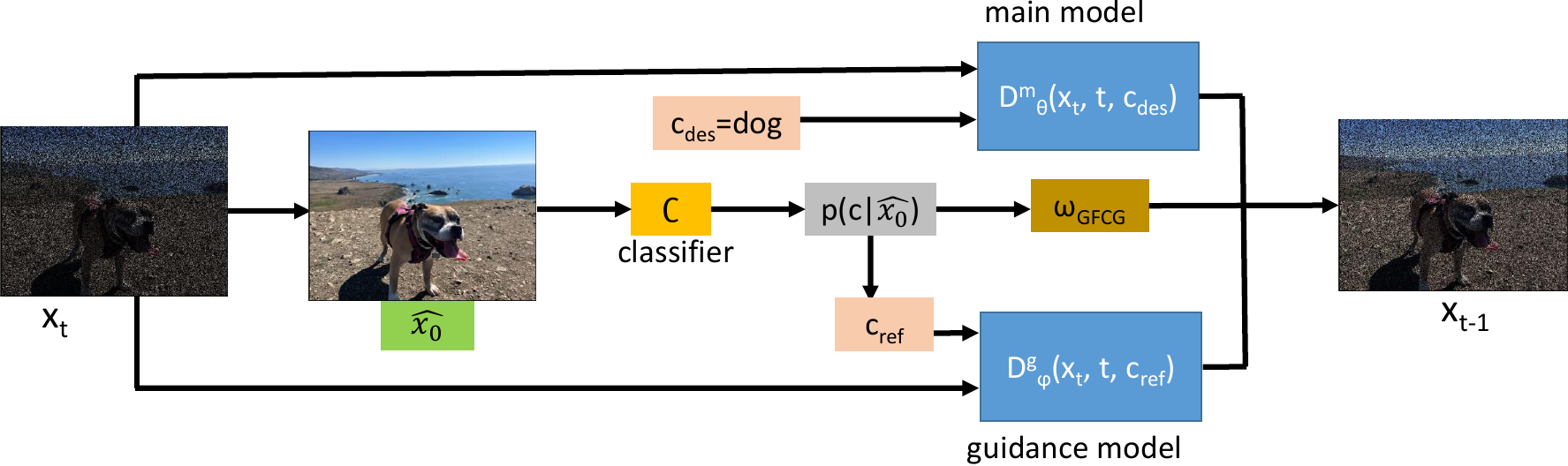}
  \caption{Schematic diagram of the proposed GFCG, which use a pre-trained classifier to guide diffusion sampling away from inaccurate class feature without time-consuming gradient descent.}
  \label{fig:diagram}
  \vspace{-10pt}
\end{figure*}

\subsection{Gradient-free Classifier Guidance}
Classifier guided sampling \cite{dhariwal_nips_2021} from diffusion models involves the computation of gradients of classifier probabilities, and this is computationally expensive
as it involves the use of autograd operators. To mitigate this issue, as explained above, CFG\cite{ho_nipsw_2021} was proposed to use an unconditional sample as a reference to increase contrast from.
We extend these concepts to devise a novel formulation that utilizes a classifier, 
without computation of gradients, to generate an conditional sample as the reference. In what follows, we describe this methodology and refer to our method as ``gradient-free
classifier guidance" (GFCG). Our method is also adaptive in that it computes the guidance scale on-the-fly depending on how {\it confused} the diffusion model is during the denoising process. 

In what follows, we will describe the method to a class-conditional diffusion model, where we denote the class label as $c_i$ for the $i$-th class (e.g., i = 0, 1, ..., 999 for ImageNet which
has 1000 classes). 
Let $c_{\text{des}}$ refer to the desired class that we wish to generate. 
At each time step $t$, the noisy sample $x_t$ is used to estimate the corresponding noise-free $\widehat{x}_{0}$:
\begin{equation}\label{eq:x0} 
\widehat{x}_{0} = \frac{x_t - \sqrt{1-\alpha_t} D^{m}_{\theta}(x_t, t, c_{\text{des}})}{\sqrt{\alpha_t}},
\end{equation}
where $\alpha_t$ is the noise schedule at time $t$ \cite{ho_nips_2020, song_iclr_2021}. A pre-trained classifier $\mathcal{C}$ is used to estimate the class probabilities:
$p(c_i|\widehat{x}_0)$. We then define an adaptive guidance scale as follows:
\begin{eqnarray}\label{eq:w}
\omega= \begin{cases}
1 + \alpha e^{-\beta (p(c_{\text{des}}|\widehat{x}_0)-\tau)} & \text{ if } \,\,\, p(c_{\text{des}}|\widehat{x}_0) < \tau \\
1 & \text{ otherwise }  
\end{cases}
\end{eqnarray}
where $\alpha>$0, $\beta>$0 and 0$\le\tau\le$1 are hyperparameters that need to be calibrated. When $p(c_{\text{des}}|\widehat{x}_0) < \tau$, the diffusion
model is {\it confused}, necessitating the use of guidance to help the data generation. On the other hand, if $p(c_{\text{des}}|\widehat{x}_0) \ge \tau$, the diffusion model
is confident and does not require guidance in this scenario. Thus, we adaptively determine if guidance is required at any time step and also the right magnitude.
In addition, if $p(c_{\text{des}}|\widehat{x}_0) < \tau$, we identify a reference class $c_{\text{ref}}$ following two criteria:
(1) $c_{\text{ref}}$ = the class with highest probability if $c_{\text{des}}$ does not have the highest probability; 
(2) $c_{\text{ref}}$ = the class with second highest probability if $c_{\text{des}}$ has the highest probability.   
We then recast the denoising step in Equation \ref{eq:g} as follows:
\iffalse
\begin{eqnarray}  
d{\bf x} = \left[{\bf f}({\bf x}, t) - g(t)^2 \widehat{D} \right] dt + g(t)d\overline{{\bf w}}; \nonumber \\ 
\widehat{D} = \omega D^{m}_{\theta}({\bf x}, t, c_{\text{des}}) - (\omega-1) D^{g}_{\phi}({\bf x}, t, c_{\text{ref}}). 
\end{eqnarray}
\fi
\begin{equation}
\widehat{D} = \omega D^{m}_{\theta}({\bf x}, t, c_{\text{des}}) - (\omega-1) D^{g}_{\phi}({\bf x}, t, c_{\text{ref}}). 
\end{equation}
Note that this involves two forward propagation processes, one with $c_{\text{des}}$ and another with $c_{\text{ref}}$ as the conditioning class.

In summary, this method adaptively determines the guidance strength using a pre-trained classifier, and also decides when to use/not use guidance. 
For the classifier $\mathcal{C}$, we use a standard ResNet-101 pre-trained on ImageNet.

\begin{algorithm}[h!]
\textbf{Input:} (i) Trained diffusion models $D^{m}_{\theta}$ and $D^{g}_{\phi}$; \\
\hspace{2.7em} (ii) Noise schedule: $\alpha_t, \overline{\alpha}_t$, $0 \le t < T$; \\
\hspace{2.5em} (iii) Hyperparameters $\alpha$, $\beta$, $\tau$, $t_s$ $\&$ $s_{cp}$; \\
\hspace{2.65em} (iv) Trained classifier, $\mathcal{C}$. \\
\textbf{Output:} A generated (noise-free) image, $x_{0}$. \\
 \vspace{1mm} \hrule \vspace{1mm}
\SetAlgoLined
 $x_{T} \sim \mathcal{N}(0, {\bf I})$ \\
 Desired class: $c_{des} \sim (c_1, c_2, \ldots, c_N)$ \\
 \For{$t=T,T-1,\ldots,1$}{
    $D_1 \leftarrow D^{m}_{\theta}(x_t, t, c_{des})$; \\
    $\widehat{D} \leftarrow D_1$;\\
    \uIf {$t \leq t_{s}$}{
        \uIf{ ($t_{s}-t)\hspace{2pt} \%\hspace{2pt} s_{cp} == 0 $}{
        \text{Calculate} $\widehat{x}_0$ \text{using Equation \ref{eq:x0}}; \\
        Use $\mathcal{C}$ to estimate $p(c|\widehat{x}_0)$; \\
        % $\omega \leftarrow 1 + \alpha e^{-\beta (p(c_{des}|\widehat{x}_0)-\tau)}$; \\ 
        \text{Evaluate} $\omega$ \text{using Equation \ref{eq:w}}; \\
        Estimate the reference class, $c_{ref}$; \\ 
        }
        \uIf {$p(c_{des}|\widehat{x}_0) < \tau$}{
            $D_2 \leftarrow D^{g}_{\phi}(x_t, t, c_{ref})$; \\
            $\widehat{D} \leftarrow \omega D_1 - (\omega-1) D_2$; \\
        }        
    } 
    Evaluate $x_{t-1}$ using Euler or Heun's solver: $x_{t-1} = \frac{1}{\sqrt{\alpha_t}}\left( x_t - \frac{1-\alpha_t}{\sqrt{1-\overline{\alpha}_t}}\widehat{D} \right)$; \\
 }
 \textbf{Return} $x_0$ 
 \caption{Gradient-free Classifier Guidance}
\label{alg:gfcg}
\end{algorithm}

\begin{figure*}[t!]
\captionsetup[subfigure]{labelformat=empty}
  \centering
  \vspace{-1pt}
  \includegraphics[width=0.85\textwidth, interpolate=false]{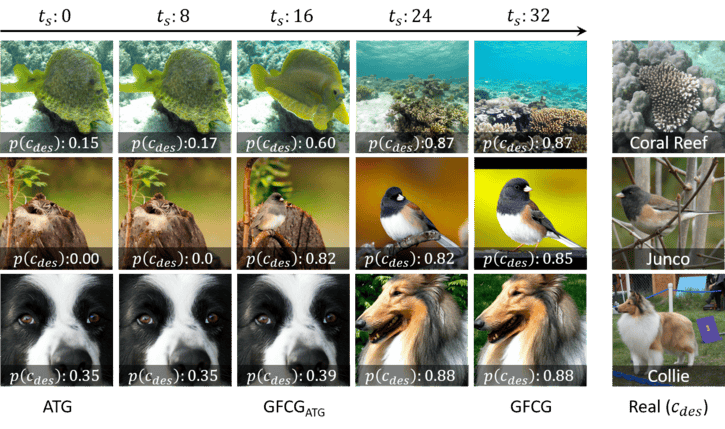} 
  \vspace{-5pt}
  \caption{Visual examples of generated ImageNet class images, combining GFCG with ATG as a function of $t_{s}$ in EDM2-XXL sampling.}
  \label{fig:edm2_v}
  \vspace{-15pt}
\end{figure*}

\subsection{Implementation Considerations}

As explained above, the proposed guidance can be applied at each sampling step using adaptively determined $c_{ref}$ and $\omega$.  In practice, there are two additional hyperparameters, $t_s$ and $s_{cp}$, which can also be tuned for optimal guidance effects.  Similar to the guidance interval applied to CFG~\cite{kynkaanniemi_arxiv_2024}, $t_s$ is the starting time step that the proposed GFCG is applied first and beyond during the denoising steps.  For $s_{cp}$, it is used to determine the frequency for classifier prediction for $c_{ref}$ and $\omega$ adaptation, \textit{i}.\textit{e}., the classifier is invoked only once every $s_{cp}$ steps. The pseudo-code to implement the GFCG guidance method is included in Algorithm \ref{alg:gfcg}.

There are also two optional techniques that can be added to improve guidance effects based on the models used.  First for $\widehat{x}_0$, in place of the one step calculation as in Equation \ref{eq:x0}, a multi-step denoising process can be used to improve classifier prediction accuracy, especially in the case when $s_{cp}$ is large, like classifier prediction is only used once.   Additionally, in place of the deterministic reference class $c_{ref}$ as explained in previous section, a stochastic reference class can be sampled at each step following Equation \ref{eq:pref} where $p_{ref}(c_i)$ is the probability that $c_i$ is chosen as $c_{ref}$.  This technique is helpful when $s_{cp}$ is small, like classifier prediction is invoked for every sampling step.
\begin{equation}\label{eq:pref} 
p_{ref}(c_i) = p(c_i|\widehat{x}_0) / \sum_{j \sim \{1, 2, \ldots, N\}-\{des\}} p(c_j|\widehat{x}_0).
\end{equation}

\section{Experiments}
\label{sec:exp}

The main experiments are conducted with the EDM2~\cite{karras_cvpr_2024} code base\footnote{https://github.com/NVlabs/edm2} and pre-trained ImageNet~\cite{deng_cvpr_2009} 512$\times$512 class-conditional models. Application of the proposed GFCG on text-to-image models is also investigated where the Stable Diffusion (SD) 1.5\footnote{https://huggingface.co/stable-diffusion-v1-5/stable-diffusion-v1-5} model is used for image sampling.  Bird Species\footnote{https://huggingface.co/datasets/chriamue/bird-species-dataset}, a fine-grained classification dataset which consists of 525 different bird species, and it's accompanying classifier\footnote{https://huggingface.co/chriamue/bird-species-classifier} are used for guided sampling and assessment of generated image quality.

To evaluate the overall sample quality for various methods compared in this work, the new FD\textsubscript{DINOv2}~\cite{stein_nips_2024} is used in favor of the original Fr\'echet Inception distance (FID)~\cite{heusel_nips_2017}.  According to \cite{stein_nips_2024}, the Inception encoder used in FID is unfairly punitive to diffusion models.  More importantly, it was noted in \cite{parmar_cvpr_2022} that FID is sensitive to minor data domain shift like resizing kernels and lossy compression qualities.  We have conducted a comparison test between FID and \FDD by compressing the same set of guided samples with different qualities, which shows \FDD is almost constant across different settings which FID is very sensitive to.  Additionally, 
we adopt the Precision and Recall metrics used in CFG to measure image fidelity and diversity separately.  The image fidelity metric Precision is computed as the percentage of generated samples that fall into the data manifold (assessed using the real image classifier), while the image diversity metric Recall is measured as the fraction of real images which is correctly classified by the classifier trained from sampled images. Recall is only reported for a subset of classes with selected experiments due to the time-consuming processes of sample generation and classifier training.

\begin{table*}[h!]
	\centering
	\footnotesize
	\setlength{\tabcolsep}{4pt}
\vspace{0pt}
	\caption{Comparison of GFCG and other gradient-free guidance methods using {\FDD} and classification metrics for 50,000 generated samples from EDM2, assessed with the ImageNet dataset. The best in each metric per section is highlighted in \textbf{bold} and the second best is marked with \underline{underline}.  The best metric of all EDM2-S tests is also highlighted in {\color{blue} \textbf{blue}}.}
\vspace{0pt}
\begin{threeparttable}
	\begin{tabular}{c|c|ccc|cccccccc} 
		\hline
		{\textbf{Model}} & {\textbf{Method}} & {\textbf{\FDD$\downarrow$}} & {\textbf{Precision}$\uparrow$} & {\textbf{Recall}$\uparrow$\tnote{a}} & {$M_g$} & {$\omega_{ATG}$} & {$\omega_{CFG}$} & {$\omega_{SEG}$} & {$\alpha$} & {$\beta$} & {$EMA_{m}$} & {$EMA_{g}$}  \\
		\hline \hline
        {EDM2-S} & {NG} & {68.29} & {90.0\%} & {71.5\%} & {-} & {-} & {-} & {-} & {-} & {-} & {0.190} & {-} \\
        \cline{2-13}
        {} & {CFG} & {51.80} & {\underline{94.9\%}} & {70.0\%} & {(XS,T)} & {-} & {1.90} & {-} & {-} & {-} & {0.085} & {0.085} \\
        {} & {SEG} & {\underline{39.45}} & {89.1\%} & {\textbf{75.0\%}} & {(XS,T/16)} & {-} & {-} & {2.20} & {-} & {-} & {0.085} & {0.165} \\
        {} & {ATG} & {\textbf{38.50}} & {90.6\%} & {\underline{74.1\%}} & {(XS,T/16)} & {2.45} & {-} & {-} & {-} & {-} & {0.120} & {0.165} \\
        {} & {GFCG} & {40.71} & {\textbf{95.4\%}} & {72.0\%} & {(XS,T/16)} & {-} & {-} & {-} & {0.50} & {1.25} & {0.085} & {0.165} \\
		\cline{2-13}
        {} & {GFCG\textsubscript{NG}} & {\underline{36.93}} & {93.3\%} & {\underline{75.1\%}} & {(XS,T/16)} & {-} & {-} & {-} & {0.80} & {1.25} & {0.085} & {0.165} \\
        {} & {GFCG\textsubscript{CFG}} & {44.17} & {\color{blue} \textbf{95.6\%}} & {68.8\%} & {(XS,T/16)} & {-} & {1.90} & {-} & {0.70} & {1.00} & {0.085} & {0.165} \\
        {} & {GFCG\textsubscript{SEG}} & {37.99} & {93.5\%} & {73.2\%} & {(XS,T/16)} & {-} & {-} & {2.20} & {0.80} & {1.25} & {0.085} & {0.165} \\
        {} & {GFCG\textsubscript{ATG}} & {\textbf{34.56}} & {\underline{93.5\%}} & {\color{blue} \textbf{75.3\%}} & {(XS,T/16)} & {2.45} & {-} & {-} & {0.80} & {1.25} & {0.085} & {0.165} \\
		\cline{2-13}
        {} & {GFCG\textsubscript{ATG}+CFG}  & {\color{blue} \textbf{33.39}} & {93.8\%} & {72.1\%} & {(XS,T/16)} & {2.45} & {1.60} & {-} & {0.40} & {1.25} & {0.085} & {0.165} \\
            \hline
        {EDM2-XXL} & {NG} & {42.58} & {90.8\%} & {-} & {-} & {-} & {-} & {-} & {-} & {-} & {0.150} & {-} \\
        {} & {CFG} & {32.74} & {\underline{93.7\%}} & {-} & {(XS,T)} & {-} & {1.70} & {-} & {-} & {-} & {0.015} & {0.015} \\
        {} & {ATG} & {\underline{24.83}} & {90.2\%} & {-} & {(M,T/3.5)} & {2.30} & {-} & {-} & {-} & {-} & {0.130} & {0.205} \\
        % {} & {GFCG} & {\underline{31.26}} & {\textbf{95.9\%}} & {(M,T/3.5)} & {-} & {-} & {-} & {0.40} & {1.25} & {0.095} & {0.205} \\
        %     \cline{2-12}
        {} & {GFCG\textsubscript{ATG}} & {\textbf{23.09}} & {\textbf{94.3\%}} & {-} & {(M,T/3.5)} & {2.10} & {-} & {-} & {0.70} & {1.25} & {0.095} & {0.205} \\
		\hline	\end{tabular}
\begin{tablenotes}
\item[a] 1000 samples per class for 10 challenging Imagewoof~\cite{fastai_github_2022} classes are used to train classifiers for Recall calculation on ImageNet training data.
\end{tablenotes}
\end{threeparttable}
\label{tab:edm2}
\vspace{-12pt}
\end{table*}

%-------------------------------------------------------------------------
\subsection{Class-Conditional Image Generation}
\label{subsec:ccig}
We implemented our proposed GFCG sampling algorithm on top of the publicly available EDM2~\cite{karras_cvpr_2024} code base and use ImageNet~\cite{deng_cvpr_2009} 512$\times$512 as the main dataset. Pre-trained models of EDM2-S and EDM2-XXL with default sampling parameters: 32 deterministic steps with $2^{nd}$ order Heun Sampler~\cite{karras2022elucidating} were utilized in this work. For the classifier, we use the standard ResNet-101 pre-trained on ImageNet.

%most experiments were conducted using
As shown in Table \ref{tab:edm2}, EDM2-S was used as the main model for most experiments.  GFCG was first compared with other gradient-free guidance methods.  ATG still enjoys the best score in \FDD but it comes with a significant cost in Precision.  Comparing to ATG, our proposed GFCG is able to increase Precision significantly with some trade-off in \FDD.  To maintain the same computational efficiency, the same guidance model as in ATG is applied for GFCG and SEG.  The second group includes mixed guidance methods which combines GFCG with other guidance methods sequentially, \textit{i}.\textit{e}.,  GFCG, following other methods, is applied to steps after $t_s$.  In the case of GFCG\textsubscript{NG}, it is similar to applying guidance interval~\cite{kynkaanniemi_arxiv_2024} to GFCG.  All mixed methods improve both \FDD and Precision metrics comparing to the corresponding method which GFCG is mixed with, and GFCG\textsubscript{ATG} gets the best \FDD out of all four.  The best \FDD of all is achieved when the additive guidance of GFCG and CFG, setting a SOTA performance of \textbf{33.39} while beating ATG in Precision by a 3.2\% margin.  Experiments in EDM2-XXL shows similar benefits of GFCG.  A subset of 10 classes are also chosen for Recall assessment, which is a good indicator for sample diversity.  Individually, SEG and ATG have the two highest values.  The mixed guidance of GFCG\textsubscript{ATG} has the best recall overall while leading ATG in both \FDD and Precision too.  While assessed only on 10 classes, it shows GFCG is able to improve image fidelity while preserving diversity.  Note that the \FDD metric for ATG is a little worse than the published value~\cite{karras_arxiv_2024}. The same random seed setting was used for all methods for fair comparison.

\begin{figure*}[t!]
  \centering
  \vspace{-1pt}
  %\captionsetup[subfigure]{font=footnotesize, labelformat=empty}
  \begin{subfigure}{0.27\linewidth}
    \includegraphics[height=4.1cm, interpolate=false]{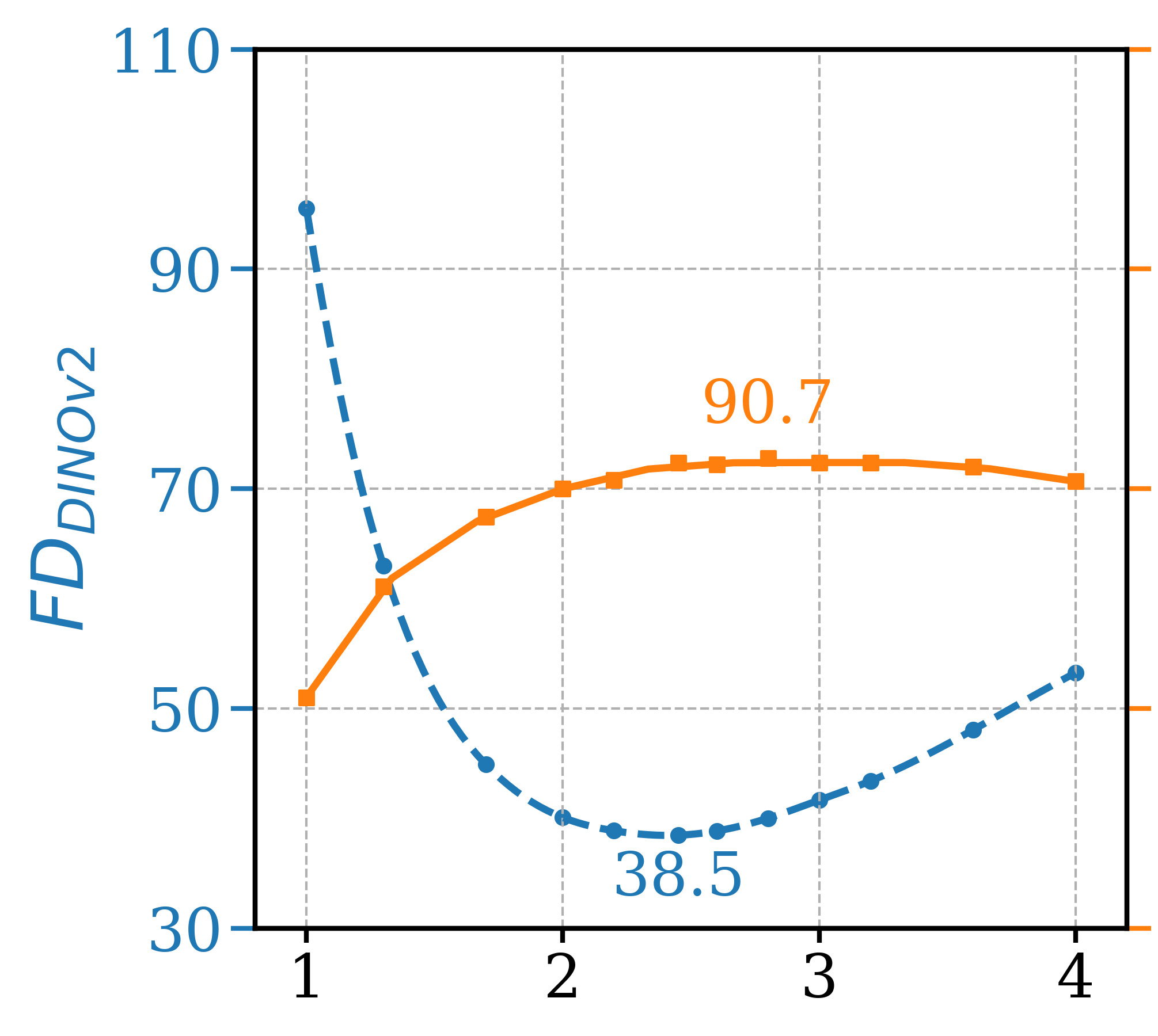}
    \caption{\hspace{0.1cm} $\omega_{ATG}$}
  \end{subfigure}
  \hfill
  \begin{subfigure}{0.22\linewidth}
    \includegraphics[height=4.1cm, interpolate=false]{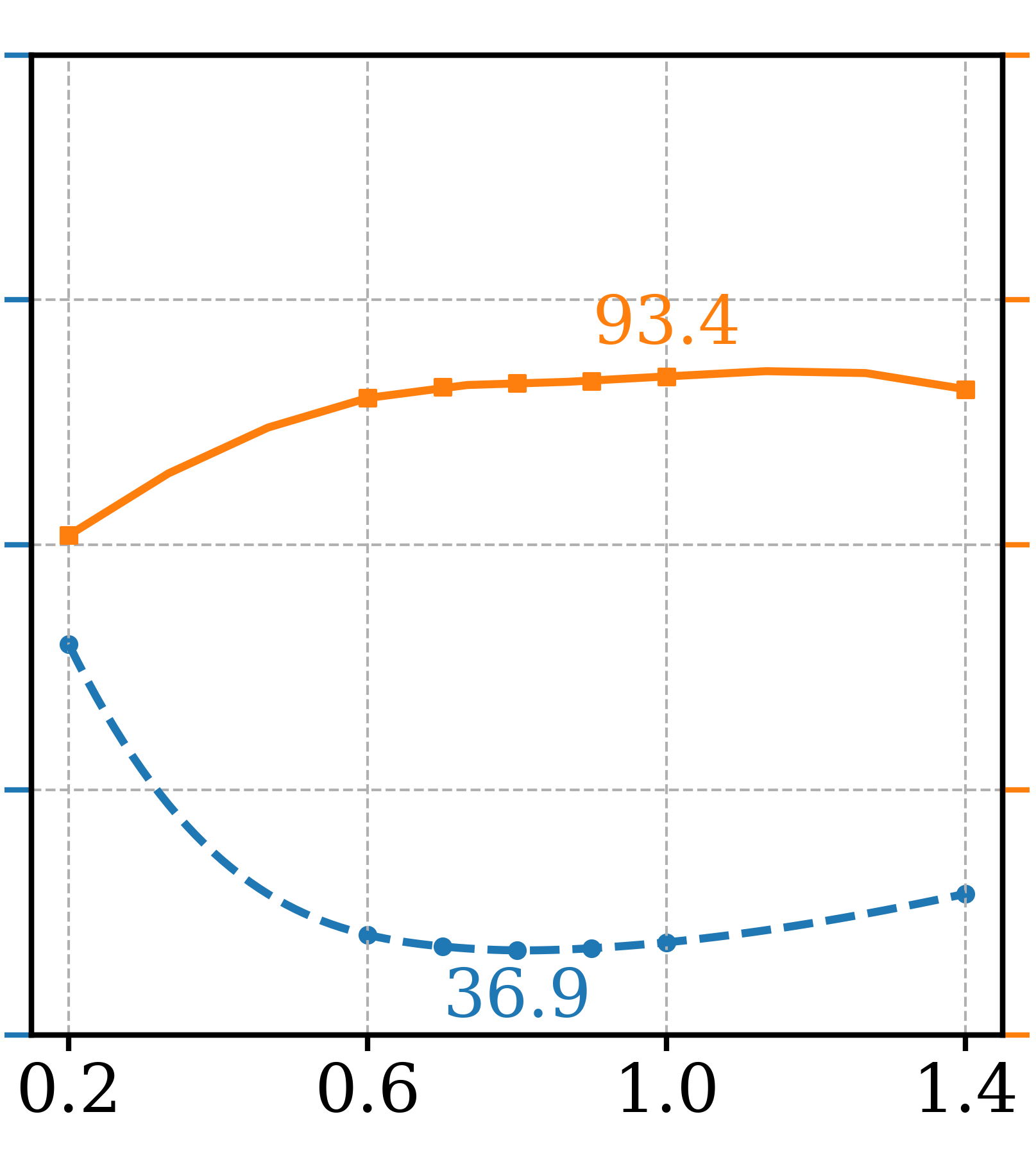}
    \caption{\hspace{0.1cm} $\alpha$}
  \end{subfigure}
  \hfill
  \begin{subfigure}{0.22\linewidth}
    \includegraphics[height=4.1cm, interpolate=false]{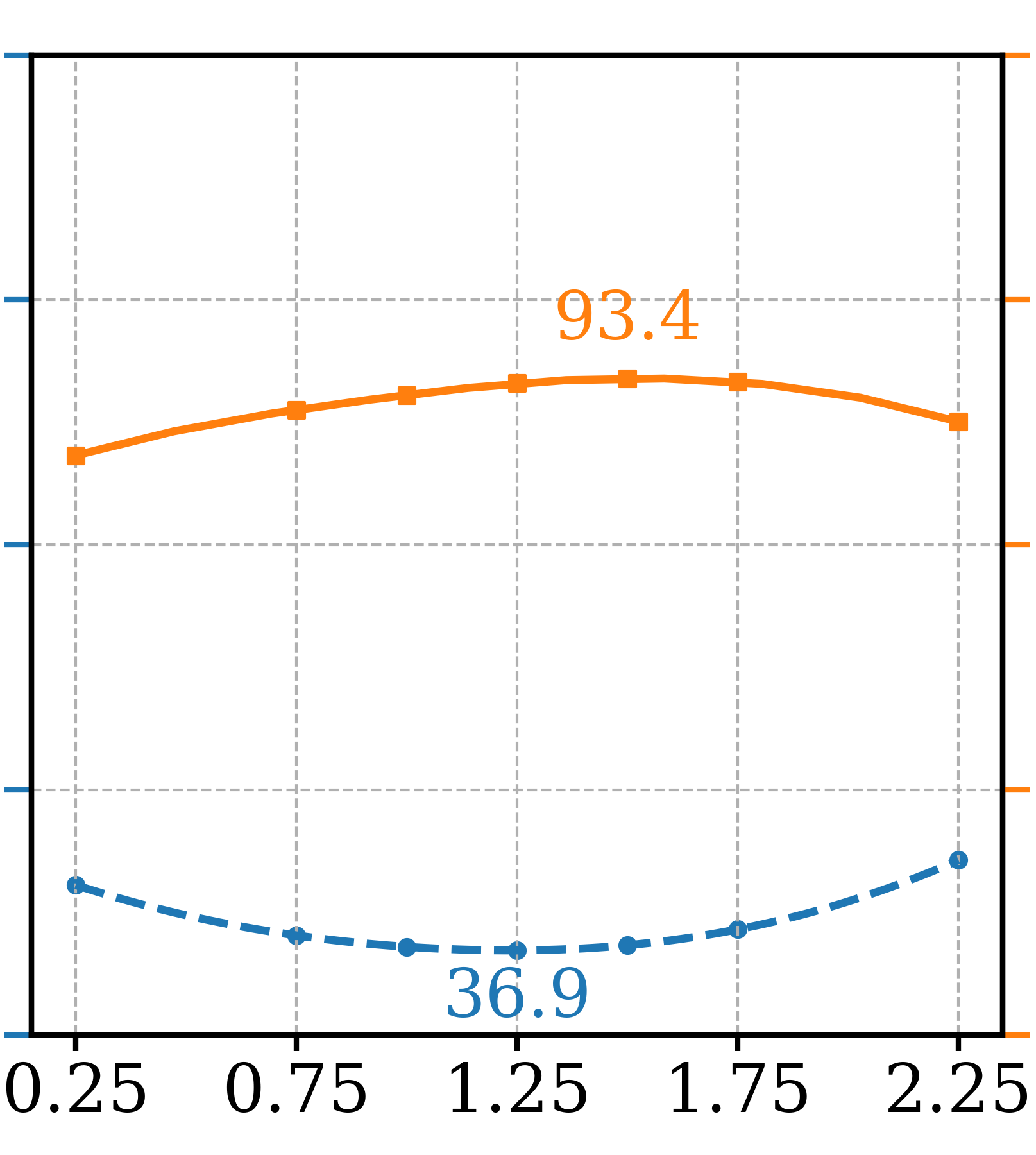}
    \caption{\hspace{0.1cm} $\beta$}
  \end{subfigure}
  \hfill
  \begin{subfigure}{0.27\linewidth}
    \includegraphics[height=4.1cm, interpolate=false]{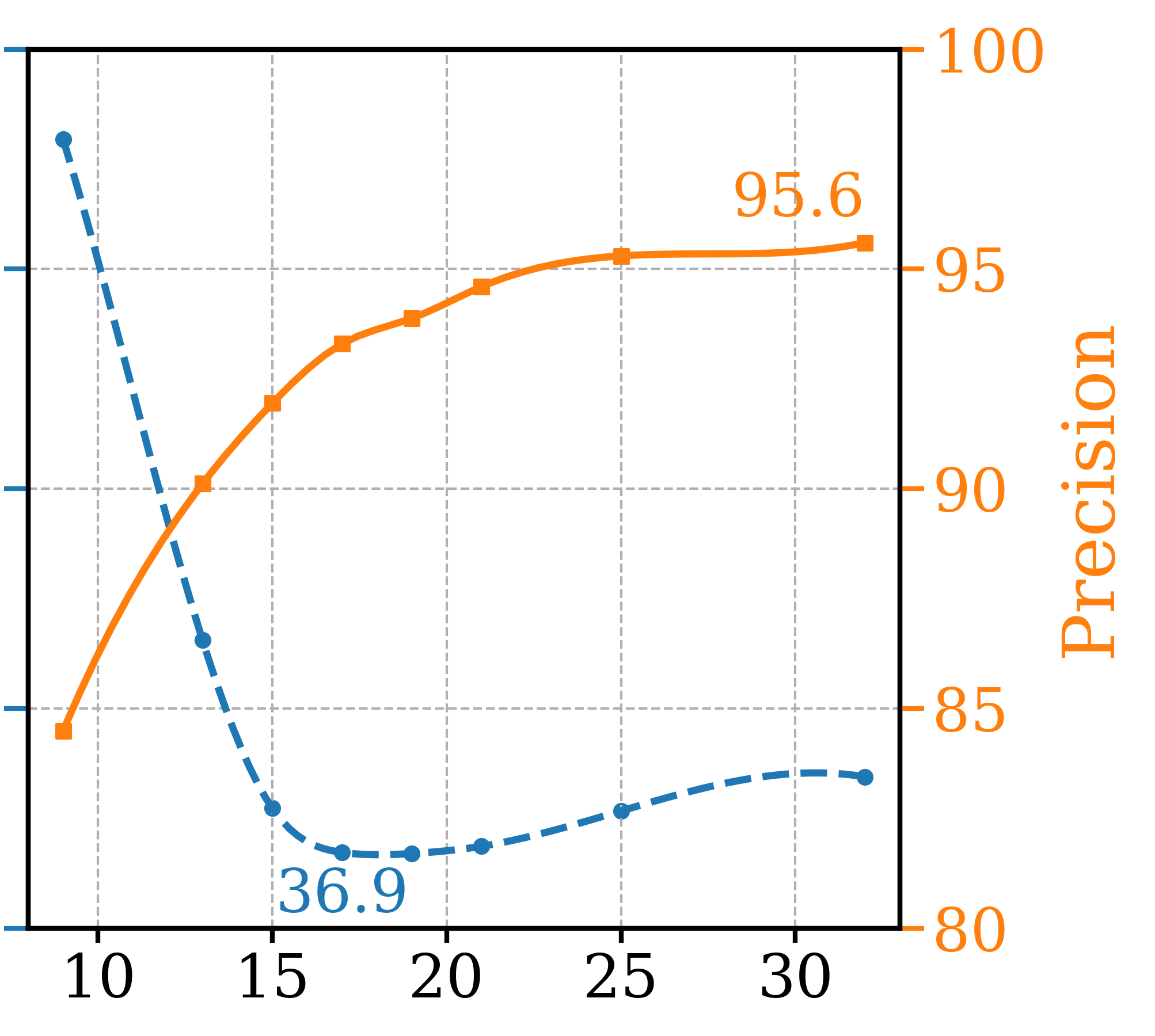}
    \caption{\hspace{0.1cm} $t_s$}
  \end{subfigure}
  \vspace{0pt}
  \caption{Ablation studies for ATG (a) and GFCG (b-d) for class-conditional image generation.}
  \label{fig:edm2_a}
  \vspace{-10pt}
\end{figure*}

To determine the optimal setting for different hyperparameters, a series of ablation tests were conducted and three most critical ones are presented in Figure \ref{fig:edm2_a}, leaving the remaining ones to the supplementary section.  Note that other than the varying parameter, other settings are fixed as GFCG\textsubscript{NG} (XS,T/16) in Table \ref{tab:edm2}.  The first is study the trade-off in image fidelity (lower Precision) for different ATG guidance scale, which is not reported in the original paper.  It shows that the optimal setting for \FDD is also in range of optimal Precision, as further increased scale doesn't increase Precision.  For GFCG tests, both $\alpha$ and $\beta$ are used to determine guidance scale $\omega$ where $\alpha$ is responsible for the overall strength while $\beta$ controls the relative strength in regards to classification confidence.  While increasing $\alpha$ improves both metrics initially, \FDD starts to get worse after $\alpha$ passes 0.8 and Precision also decreases after 1.2.  For $\beta$, there is a similar trend and the optimal range is around 1.25 to 1.50.  Lastly, out of the total of 32 sampling steps, earlier the application of guidance (larger $t_s$) is better for Precision, but there is a trade-off for worse \FDD. The optimal range is around 17.  For $s_{cp}$, it is better set to the maximum so that only one classifier prediction is used for all mixed and additive GFCG methods included in Table~\ref{tab:edm2}.  Based on that, a multi-step denoising process is used for $\widehat{x}_0$ estimation.  This adds a few NFEs but not significant as the multi-step estimation is needed one time.  For example, for GFCG\textsubscript{ATG}, it is a 4-step estimation.

Visual examples are included in Figure \ref{fig:edm2_v} to demonstrate the effects of transitioning from ATG to GFCG, where intermediate results are from mixed guidance of ATG and GFCG, before and after $t_s$ respectively.  It shows image fidelity ($p(c_{des})$) increases when GFCG starts earlier.  In the second example, GFCG has better $p(c_{des})$ but looses details in the background, a concern of trade-off in diversity.  The mixed guidance for $t_s=16$ results in increased $p(c_{des})$ while preserving the background details.

%-------------------------------------------------------------------------
\subsection{Text-to-Image Generation}
While the proposed method fits well with class-conditional diffusion models naturally as it uses classifier for guided sampling, it can also help to improve sample quality for general text-to-image diffusion models like SD 1.5~\cite{sd_github_2022}.  As explained earlier, a classifier trained from the Bird Species dataset was used for guided sampling and the samples were evaluated against the same training dataset.  As the output resolution for SD 1.5 is 512$\times$512, they were resized to the same resolution of 224$\times$224 of the dataset before assessed for \FDD, Precision and Recall metrics.  All samples were generated in 50 steps using PNDM~\cite{liu_arxiv_2022}.  This fine-grained generation task is challenging as many of the bird species are long-tailed classes in SD 1.5.  The only other known generation test for such fine-grained classes was reported in the unified training-free classifier guidance (TFG) work~\cite{ye_arxiv_2024}.  Only unconditional generation using gradient classifier guidance was investigated and the best Precision score is only 2.24\%.  In this work, we focus on conditional generation using different guidance methods.  For a given target bird species [bird \#1], generic prompts like $\textit{a close up bird photo, [bird \#1]}$ were used to generate diverse images of said species.  In the case of GFCG, when a different bird species [bird \#2] was identified as the reference class, the [bird \#1] phrase in the target prompt was replaced by [bird \#2] to get $c_{ref}$.

\begin{figure*}[t!]
\captionsetup[subfigure]{font=footnotesize, labelformat=empty}
 \begin{center}
%  \begin{subfigure}[b]{0.96\textwidth}
%    \centering
%      \includegraphics[width=\textwidth, interpolate=false]{CVPR2025/images/53_3_2.png}
%  \end{subfigure}
%  \hspace*{0.2em}
%  \rotatebox[origin=l]{90}{\makebox[0.12\textwidth]{$\scriptstyle BALD EAGLE$}}
%
  \begin{subfigure}[b]{0.96\textwidth}
    \centering
      \includegraphics[width=\textwidth, interpolate=false]{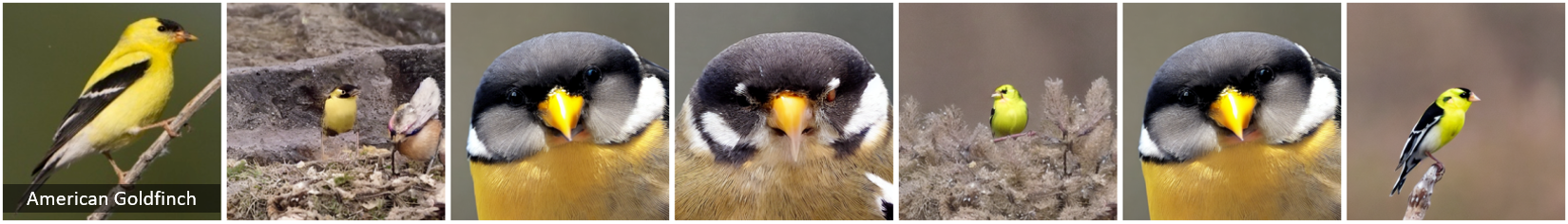}
  \end{subfigure}
  % \hspace*{0.1em}
  % \rotatebox[origin=l]{90}{\scalebox{0.75}{$\scriptstyle \hspace{8pt} american \hspace{2pt} goldfinch$}}

  \begin{subfigure}[b]{0.96\textwidth}
    \centering
      \vspace{-0.1em}
      \includegraphics[width=\textwidth, interpolate=false]{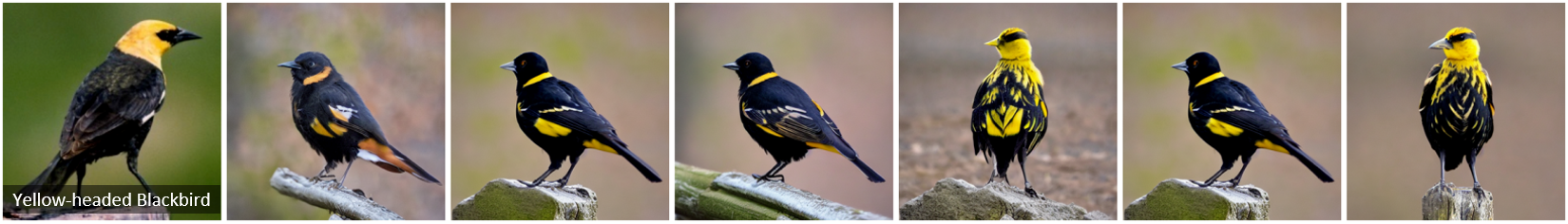}
  \end{subfigure}
% \hspace*{0.1em}
%  \rotatebox[origin=l]{90}{\makebox[0.12\textwidth]{$\scriptstyle YELLOW HEADED BLACKBIRD$}}
% \rotatebox[origin=l]{90}{\scalebox{0.75}{$\scriptstyle yellow \hspace{1pt} headed \hspace {1pt} blackbird$}}

%   \begin{subfigure}[b]{0.96\textwidth}
%     \centering
%       \includegraphics[width=\textwidth, interpolate=false]{CVPR2025/images/425_0_0.png}
%   \end{subfigure}
%   \hspace*{0.2em}
%   \rotatebox[origin=l]{90}{\makebox[0.12\textwidth]{$\scriptstyle ROSEATE SPOONBILL$}}

  \begin{subfigure}[b]{0.96\textwidth}
    \centering
      \vspace{-0.1em}
      \includegraphics[width=\textwidth, interpolate=false]{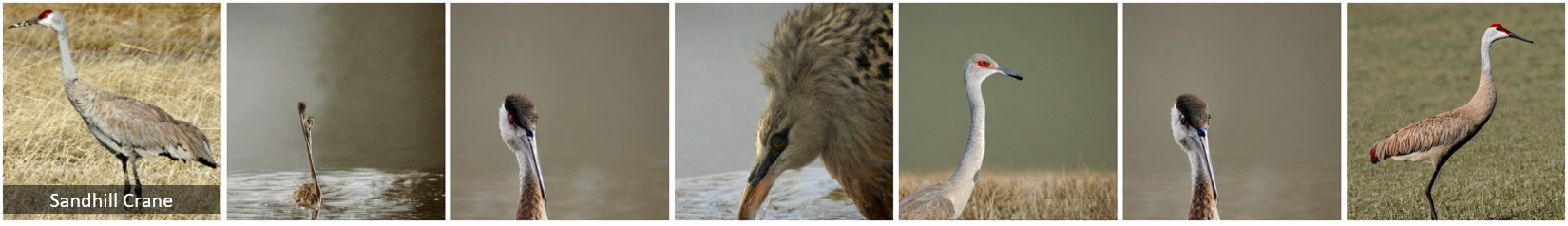}
      \caption{\hspace{0em} Real \hspace{6.5em} NG \hspace{7em} CFG \hspace{6em} PAG \hspace{6em} GFCG \hspace{4.5em} GFCG\textsubscript{CFG} \hspace{3.5em} GFCG+CFG}
  \end{subfigure}
  % \hspace*{0.1em}
  % \rotatebox[origin=l]{90}{\scalebox{0.75}{$\scriptstyle \hspace{30pt} sandhill \hspace{2pt} crane$}}
 \end{center}
 \vspace{-10pt}
 \caption{Visual examples of generated fine-grained class images from SD 1.5 using GFCG and other gradient-free guidance methods.}
 \label{fig:sd15_v}
\end{figure*}

%Optimal settings were selected for all methods based on the \FDD and Precision trade-off in ablation studies.  
GFCG was first compared with three gradient-free guidance methods: CFG, PAG and SEG.  ATG is not included as training a bad version of SD 1.5 with the optimal settings is not trivial.  For experiment results shown in Table \ref{tab:birds}, 80 samples were generate for each of the 525 species and the same random seed settings were kept across different methods for fair comparison.  Among all four, CFG has the best \FDD metric while our GFCG has a clear advantage in Precision score.  PAG and SEG don't perform well for this challenging task, likely due to lack of reliable self-attention weights for long-tailed classes.  Experiments were also conducted for mixed guidance GFCG\textsubscript{CFG}, where CFG was used exclusively before applying GFCG after $t_s=20$.  Without additional NFEs, it is able to match \FDD of CFG while maintaining advantage in Precision.  The best results are additive guidance of GFCG+CFG, both \FDD and Precision metrics are improved significantly though at the cost of double NFEs.  Comparing to ImageNet experiments where it is beneficial to start applying GFCG midway through sampling and predict confused class only once, here it was consistently found advantageous to start applying GFCG from the beginning and conduct classifier prediction more frequently.  Consequently, the stochastic selection of $c_{ref}$ is also included to improve performance.  This is likely due to the challenging case of fine-grained classification as there are multiple confused classes, so that adjusting confused class using classifier prediction more often is desired.
As SD 1.5 is trained on a large general dataset, GFCG based on the Bird Species classifier can only provide guidance inside the bird related region of the overall data distribution, while CFG adds guidance from other parts of the data distribution.  This could be the cause that GFCG+CFG has the best in both metrics.

For visual examples in Figure \ref{fig:sd15_v}, the benefits of GFCG is obvious in terms of image fidelity, like the yellow head in the second one.  The mixed guidance of GFCG\textsubscript{CFG} keeps the general structure as in CFG, but improves image details which results in improved Precision comparing to CFG.  The best results are from additive guidance of GFCG+CFG, which improves the image fidelity in both overall composition as well as fine details.

\begin{figure*}[t!]
  \centering
  \vspace{-1pt}
  %\captionsetup[subfigure]{font=footnotesize, labelformat=empty}
  \begin{subfigure}{0.27\linewidth}
    \includegraphics[height=4.1cm, interpolate=false]{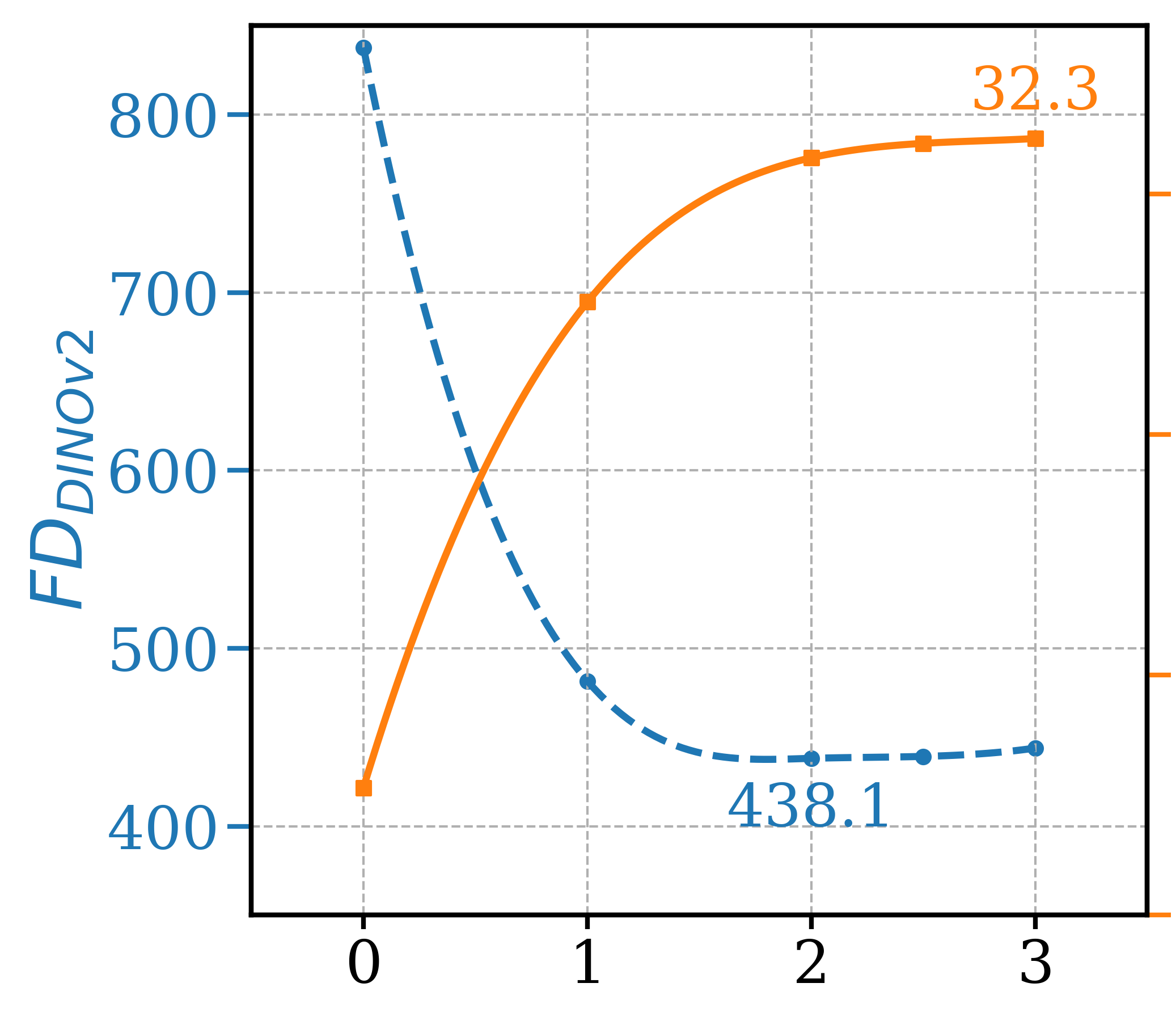}
    \caption{\hspace{0.1cm} $\alpha$ (GFCG)}
  \end{subfigure}
  \hfill
  \begin{subfigure}{0.22\linewidth}
    \includegraphics[height=4.1cm, interpolate=false]{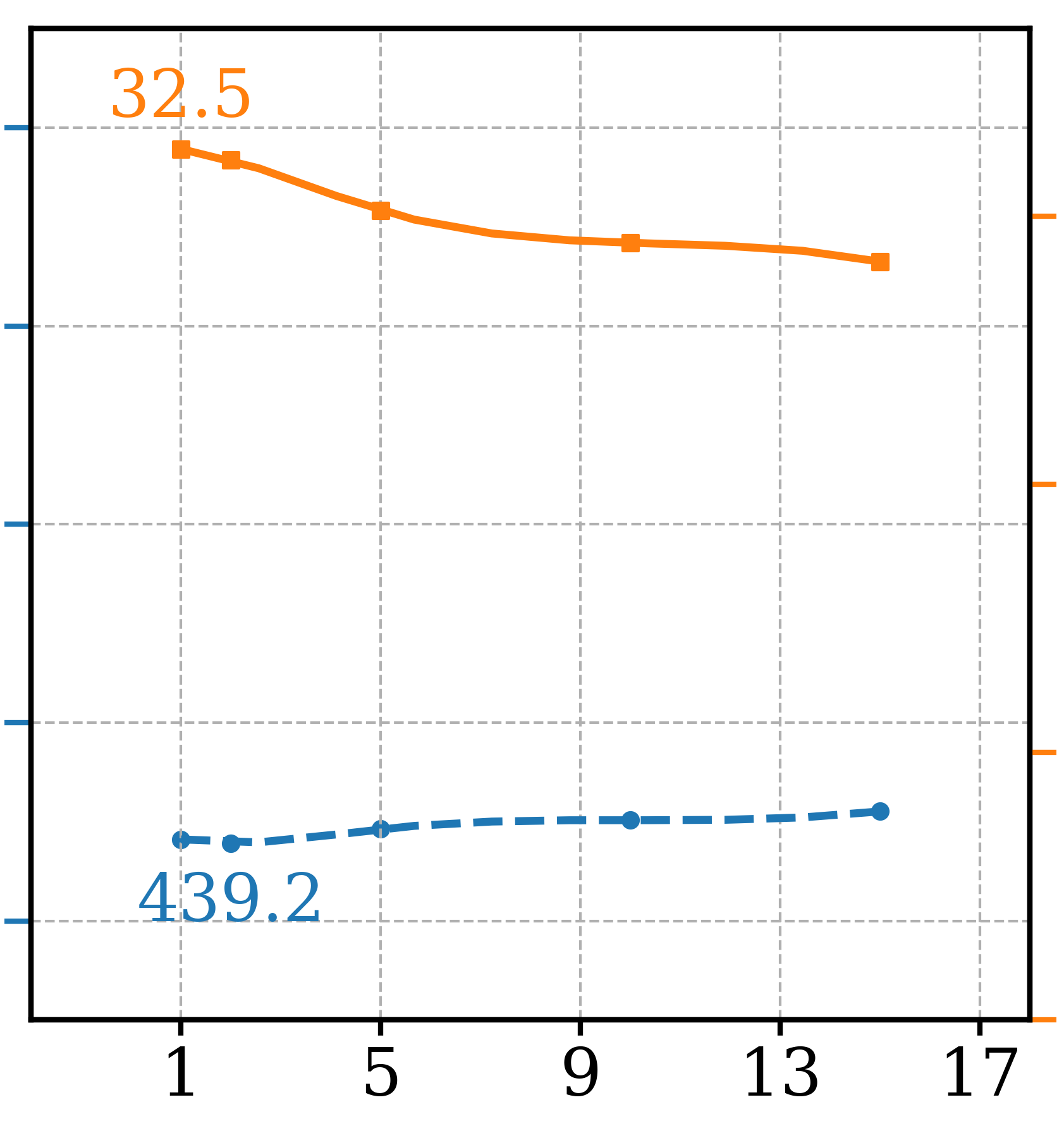}
    \caption{\hspace{0.1cm} $s_{cp}$ (GFCG)}
  \end{subfigure}
  \hfill
  \begin{subfigure}{0.22\linewidth}
    \includegraphics[height=4.1cm, interpolate=false]{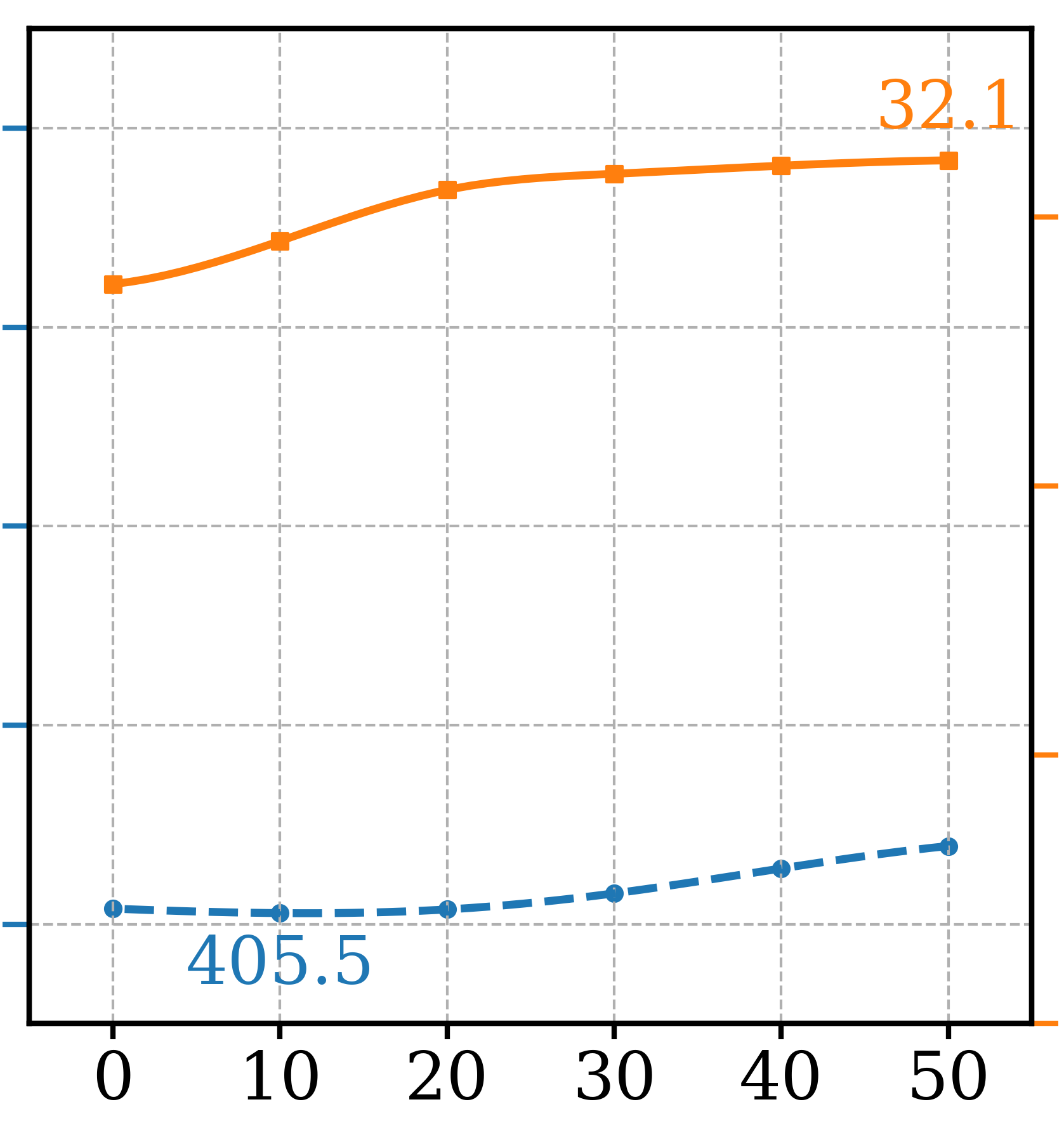}
    \caption{\hspace{0.1cm} $t_s$ (GFCG\textsubscript{CFG})}
  \end{subfigure}
  \hfill
  \begin{subfigure}{0.27\linewidth}
    \includegraphics[height=4.1cm, interpolate=false]{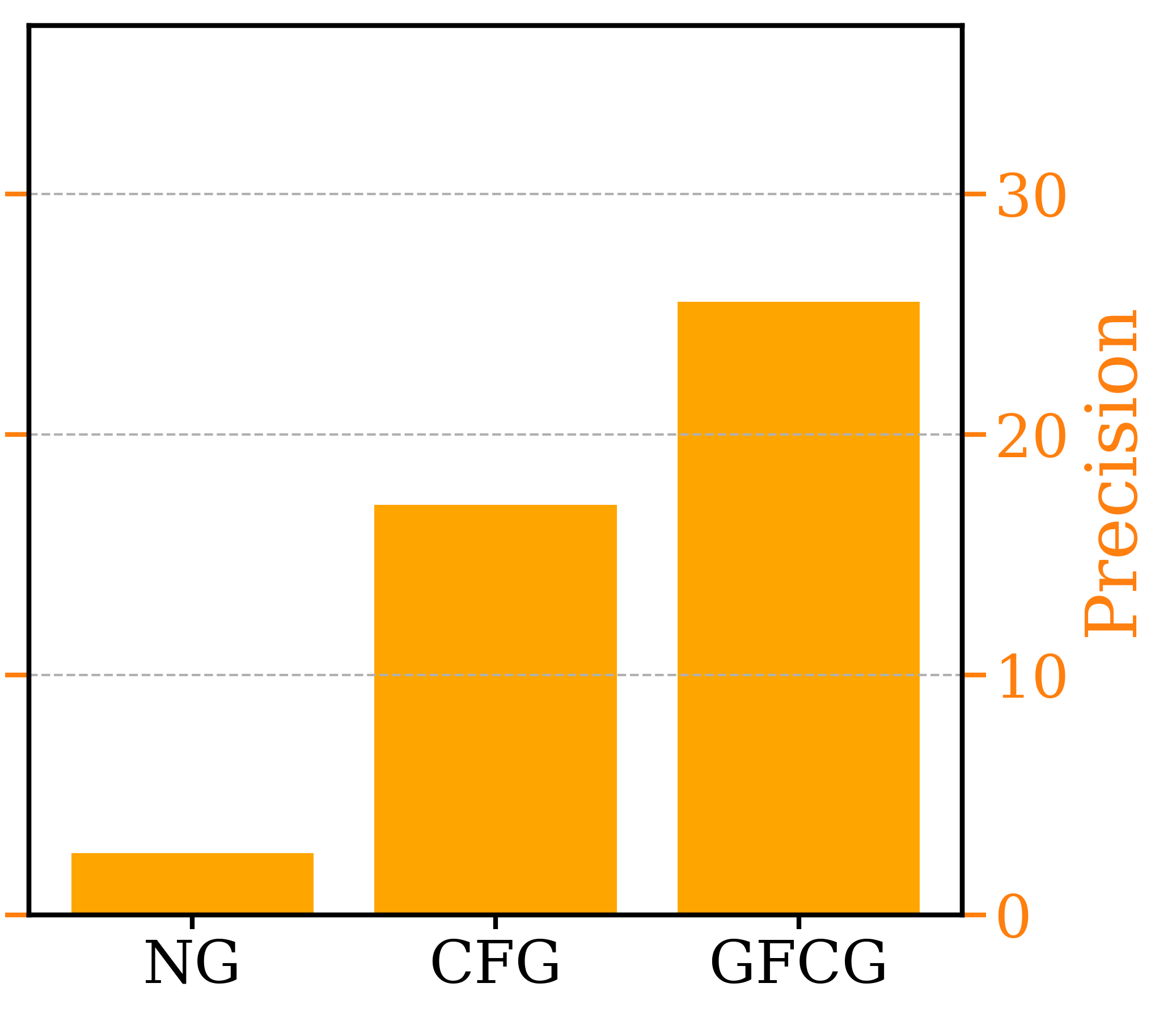}
    \caption{\hspace{0.1cm} guidance methods}
  \label{fig:sd15_ad}
  \end{subfigure}
  \vspace{0pt}
  \caption{(a-c): Ablation studies for text-to-image generations (8,400 samples); (d): Experiments of detailed prompts (21,000 samples).}
  \label{fig:sd15_a}
  \vspace{-10pt}
\end{figure*}

% \begin{figure}[h!]
%   \centering
%   \includegraphics[width=\linewidth]{CVPR2025/images/sd15-crop.pdf}
%   \caption{Ablation studies for optimal settings of GFCG methods.}
%   \label{fig:sd15_a}
% \end{figure}

\begin{table}[h!]
	\centering
	\footnotesize
	\setlength{\tabcolsep}{2.4pt}
\vspace{0pt}
	\caption{Comparison of GFCG and other gradient-free guidance methods using 42,000 generated samples from SD 1.5, assessed with the Bird Species dataset.  For the five methods with the same NFEs, the best in each metric is highlighted in \textbf{bold} and the second best is marked with \underline{underline}. The best metric of all methods is also highlighted in {\color{blue} \textbf{blue}}.}
\vspace{1pt}
	\begin{tabular}{r|cc|cccccc} 
		\hline
		{} & {\FDD$\downarrow$} & {Precision$\uparrow$} & {$\omega_{CFG}$} & {$\omega_{PAG}$} & {$\omega_{SEG}$} & {$\alpha$} & {$t_s$} \\
		\hline \hline
        {NG} & {816.3} & {5.3\%} & {1.0} & {-} & {-} & {-} & {-} \\
		\hline
        {CFG} & {\underline{394.0}} & {27.3\%} & {5.5} & {-} & {-} & {-} & {-} \\
        {PAG} & {562.3} & {12.5\%} & {-} & {3.5} & {-} & {-} & {-} \\
        {SEG} & {676.4} & {8.0\%} & {-} & {-} & {10.0} & {-} & {-} \\
        {GFCG} & {418.8} & {\textbf{32.3\%}} & {-} & {-} & {-} & {2.5} & {50} \\
        {GFCG\textsubscript{CFG}} & {\textbf{392.2}} & {\underline{30.2\%}} & {5.5} & {-} & {-} & {2.5} & {20} \\
		\hline
        {GFCG+CFG} & {\color{blue} \textbf{377.6}} & {\color{blue} \textbf{32.4\%}} & {4.0} & {-} & {-} & {1.5} & {50} \\
		\hline	\end{tabular}
\label{tab:birds}
\vspace{-9pt}
\end{table}

A set of ablation studies were conducted to determine the optimal settings for GFCG based sampling and three are shown in Figure \ref{fig:sd15_a}.  For GFCG, increasing $\alpha$ improves Precision consistently but \FDD starts to worsen when it is beyond 2. For $s_{cp}$, from every 10 to 2 steps, there is significant gain in Precision with \FDD also improved.  For the mixed guidance of GFCG\textsubscript{CFG}, where GFCG is applied after time step $t_s$, later application of GFCG improves \FDD at a cost of lower Precision.  Two other settings, $\beta$ and $\tau$, were both kept as 1 without fine-tuning.  Note that similar ablation studies were also conducted for compared methods like CFG, PAG and SEG for fair comparisons in Table \ref{tab:birds} and results are included in the supplementary section.  As shown in Figure \ref{fig:sd15_ad}, a set of detailed prompts, like the one in Figure \ref{fig:front}, were also designed to compare different guided sampling for their ability to maintain classification accuracy.  In this challenging case, the advantage of GFCG is more significant than the main tests of generic prompts.

\section{Conclusions and Limitations}

In this work, we have shown that our method generates high fidelity images without incurring the additional computational costs associated with classifier guidance or requiring the training of an extra unconditional model, as seen with classifier-free guidance. By leveraging an off-the-shelf classifier, GFCG can be seamlessly integrated with existing sampling methods without additional NFEs, thereby enhancing both \FDD and Precision metrics without loss in Recall.
Moreover, we further demonstrated that it can be extended to text-to-image diffusion models, achieving high class accuracy, particularly for long-tailed and fine-grained classes. This flexibility and efficiency make GFCG a robust and adaptable approach, offering significant improvements without extra training efforts or computational resources.  These results underscore the potential of GFCG to optimize overall image quality, Precision in particular, across various models and applications, paving the way for further innovations in image generation and related fields.

\noindent\textbf{Limitations:} While it is applicable to general text-to-image models which learn from a huge dataset, the proposed GFCG is most beneficial to generation of images which are in distribution of the pre-trained classifier. 
 For the Bird Species experiments, as the classifier training dataset consists of close-up shots mostly, guided sampling of bird images in other layouts would be less effective. \textbf{Ethical considerations:} We acknowledge that this guidance method could potentially benefit creation of inappropriate materials.  Deployments of such methods should apply appropriate safeguards to prevent malicious and illegal uses.

{
    \small
    \bibliographystyle{ieeenat_fullname}
    \bibliography{main,mybib_2411}
}

% WARNING: do not forget to delete the supplementary pages from your submission 
\clearpage
\setcounter{page}{1}
\maketitlesupplementary

%-------------------------------------------------------------------------
\section{Evaluation Metrics}
We have explained the reasoning for choosing \FDD over FID as the overall image quality metric in the main paper.  To further validate this choice, we also conducted a lossy compression test as in \cite{parmar_cvpr_2022} to compare FID and \FDD.  As shown in Table \ref{tab:metric}, For the same 42000 images sampled from SD 1.5, comparing to the original uncompressed output, multiple JPEG compressions are applied with different quality settings.  For \FDD, the metric remains relatively consistent for original ouput as well as different compression.  In contrast, as the Bird Species dataset used for assessment consists of JPEG images, FID benefits from applying similar JPEG compression to the original outputs.  In fact, FID improves (lower value) continuously with loss of image quality until it reaches the lowest around 80\% quality.  Lastly, it is shown in Autoguidance~\cite{karras_arxiv_2024} that the model and sample settings need to be tuned for optimal FID and \FDD separately.  As a result, \FDD is chosen as the primary sample quality metric for experiments in this work to determine optimal settings.

\begin{table}[h!]
	\centering
	\footnotesize
	\setlength{\tabcolsep}{3pt}
\vspace{0pt}
	\caption{Comparison of different image compression qualities using FID and {\FDD} metrics for 42000 generated samples from SD 1.5 when assessed using the Bird Species dataset.}
\vspace{1pt}
	\begin{tabular}{r|ccccccc} 
		\hline
		{} & {Original} & \multicolumn{5}{c}{JPEG Compression Quality} \\
		{} & {Output} & {100\%} & {95\%} & {90\%} & {85\%} & {80\%} & {75\%} \\
		\hline \hline

        {FID$\downarrow$} & {13.29} & {14.20} & {12.87} & {10.57} & {6.48} & {6.47} & {6.52} \\
        {\FDD$\downarrow$} & {401.5} & {400.5} & {399.2} & {397.0} & {397.3} & {397.6} & {398.0} \\
		\hline	\end{tabular}
\label{tab:metric}
\vspace{-9pt}
\end{table}

\section{Additional Implementation Details}
\subsection{Class-Conditional Generation: Pseudo Code}

\begin{algorithm}[h!]
\textbf{Input:} (i) Trained diffusion models $D^{m}_{\theta}$ and $D^{g}_{\phi}$; \\
\hspace{2.7em} (ii) Noise schedule: maximum noise $\sigma_t$; \\
\hspace{2.5em} (iii) Hyperparameters $\alpha$, $\beta$, $\tau$, $t_s$ $\&$ $s_{cp}$; \\
\hspace{2.65em} (iv) Trained classifier, $\mathcal{C}$. \\
\textbf{Output:} A generated (noise-free) image, $x_{0}$. \\
 \vspace{1mm} \hrule \vspace{1mm}
\SetAlgoLined
 $x_{T} \sim \mathcal{N}(0, {\sigma_T^2 \bf I})$ \\
 Desired class: $c_{des} \sim (c_1, c_2, \ldots, c_N)$ \\
 \For{$t=T,T-1,\ldots,1$}{
    $D_1 \leftarrow D^{m}_{\theta}(x_t, t, c_{des})$; \\
    $\widehat{D} \leftarrow D_1$;\\
    \uIf {$t \leq t_{s}$}{
        \uIf{ ($t_{s}-t)\hspace{2pt} \%\hspace{2pt} s_{cp} == 0 $}{
        \text{Calculate} $\widehat{x}_0$; \\
        Use $\mathcal{C}$ to estimate $p(c|\widehat{x}_0)$; \\
        \text{Evaluate} $\omega$ \text{using Equation \ref{eq:w}}; \\
        Estimate the reference class, $c_{ref}$; \\ 
        }
        \uIf {$p(c_{des}|\widehat{x}_0) < \tau$}{
            $D_2 \leftarrow D^{g}_{\phi}(x_t, t, c_{ref})$; \\
            $\widehat{D} \leftarrow \omega D_1 - (\omega-1) D_2$; \\
        }        
    } 
    Sample $x_{t-1}$ as $S(\widehat{D}, x_t, \sigma_{t-1})$; \\
 }
 \textbf{Return} $x_0$ 
 \caption{GFCG with Generalized Sampling Method}
\label{alg:gfcg0}
\end{algorithm}

\begin{algorithm*}[t!]
\textbf{Input:} (i) Trained diffusion models $D^{m}_{\theta}$ and $D^{g}_{\phi}$; \\
\hspace{2.7em} (ii) Base Sampling Method $M_{B} \in \{\text{NG, CFG, SEG, ATG}\}$; \\
\hspace{2.7em} (iii) Noise schedule $\sigma_{t}$ calculated using Equation \ref{eq:sigma} with parameters $(\bm{\sigma}_{min},\bm{\sigma}_{max},\rho,T)$; \\
\hspace{2.6em} (iv) Parameters for $\widehat{x}_{0}$ estimation using multi-step denoising $(\bm{\sigma}_{min}',\rho',T')$; \\
\hspace{2.6em} (v) Hyperparameters $\alpha$, $\beta$, $\tau$, $t_s$ $\&$ $s_{cp}$; \\
\hspace{2.65em} (vi) Trained classifier, $\mathcal{C}$. \\
\textbf{Output:} A generated (noise-free) image, $x_{0}$. \\
 \vspace{1mm} \hrule \vspace{1mm}
\SetAlgoLined
 $x_{T} \sim \mathcal{N}(0, \sigma^{2}_T{\bf I})$ \\
 Desired class: $c_{des} \sim (c_1, c_2, \ldots, c_N)$ \\
 \For{$t=T,T-1,\ldots,1$}{
    % $D_1 \leftarrow D^{m}_{\theta}(x_i, t_i, c_{des})$; \\
    % $\widehat{D} \leftarrow D_1$;\\
    $use_{\text{BASE}} \leftarrow \text{TRUE}$; \\
    \uIf {$t \le t_{s}$}{
        \uIf{ ($t_{s}-t)\hspace{2pt} \%\hspace{2pt} s_{cp} == 0 $}{
        $\bm{\sigma}_{max}' \leftarrow \sigma_{t}$; \\
        \text{Compute noise schedule} $\sigma'_{t'}$ \text{with parameters} $(\bm{\sigma}_{min}',\bm{\sigma}_{max}',\rho',T')$ using Equation \ref{eq:sigma}; \\
        $\tilde{x}_{T'} \leftarrow x_{t}$; \\
        \For{$t'=T',\ldots,1$}{
            \text{Evaluate $\tilde{x}_{t'-1} $ using Heun's solver and $\widehat{D}'$ computed based on $M_{B}$ (refer Algorithm 1 in \cite{karras2022elucidating})}; \\ 
            }
        $\widehat{x}_{0} \leftarrow \tilde{x}_0$ \text{(consistent with Algorithm \ref{alg:gfcg} in main paper)}; \\
        Use $\mathcal{C}$ to estimate $p(c|\widehat{x}_0)$; \\
        % $\omega \leftarrow 1 + \alpha e^{-\beta (p(c_{des}|\widehat{x}_0)-\tau)}$; \\ 
        \text{Evaluate} $\omega$ \text{using Equation \ref{eq:w}}; \\
        Estimate the reference class, $c_{ref}$; \\ 
        }
        \uIf {$p(c_{des}|\widehat{x}_0) < \tau$}{
            $D_1 \leftarrow D^{m}_{\theta}(x_t, \sigma_t, c_{des})$; \\
            $D_2 \leftarrow D^{g}_{\phi}(x_t, \sigma_t, c_{ref})$; \\
            $\widehat{D} \leftarrow \omega D_1 - (\omega-1) D_2$; \\
            $use_{\text{BASE}} \leftarrow \text{FALSE}$; \\
        }        
    } 
    \uIf {$use_{\text{BASE}}\;$}{
        \text{Compute $\widehat{D}$ based on $M_{B}$}; \\
        \text{Example: if $M_{B} = \text{ATG}$ then} \\
        \hspace{1.5em} \text{$D_1 \leftarrow D^{m}_{\theta}(x_t, \sigma_t, c_{des})$, $D_2 \leftarrow D^{g}_{\phi}(x_t, \sigma_t, c_{des})$ and $\widehat{D} \leftarrow \omega_{ATG} D_1 - (\omega_{ATG}-1) D_2$}; \\
    }
    \text{Evaluate $x_{t-1}$ using Heun's solver and $\widehat{D}$ (refer Algorithm 1 in \cite{karras2022elucidating})}; \\
 }
 \textbf{return} $x_0$ 
 \caption{Gradient-free Classifier Guidance for EDM2 with Multi-step Denoising for $\widehat{x}_0$}
\label{alg:gfcg1}
\end{algorithm*}

\noindent For Algorithm \ref{alg:gfcg}, specifics like timestep $t$, noise schedule and sampling method are illustrated using DDIM as the example.  In implementations, GFCG is applicable to different models and sampling methods and we've conducted class-conditional experiments using EDM2 with 2nd order Heun's solver and text-to-image ones using SD 1.5 with PNDM sampling.  Besides, some implementation details are omitted in Algorithm \ref{alg:gfcg}, including multi-step denoising for $\widehat{x}_0$ and mixed guidance of GFCG and other base methods like ATG.  To present the detailed pseudo code for class-conditional generation, we first generalize Algorithm \ref{alg:gfcg} to be compatible for both EDM2 and SD 1.5 experiments, as shown in Algorithm \ref{alg:gfcg0}.  In the case of DDIM sampling, $\sigma_T = 1$ is in line 6, Equation \ref{eq:x0} is used for line 4 and the following equation is used for line 20:
\begin{equation}\label{eq:xt}
x_{t-1} = \frac{1}{\sqrt{\alpha_t}}\left( x_t - \frac{1-\alpha_t}{\sqrt{1-\overline{\alpha}_t}}\widehat{D} \right)
\end{equation}

For our class-conditional experiments, we used the 2\textsuperscript{nd} order deterministic sampler from EDM (i.e., Algorithm 1 in \cite{karras2022elucidating}) in all experiments with $\bm{\sigma}(\mathbf{t})=\mathbf{t}$ and $s(\mathbf{t})=1$. We used the default settings $\bm{\sigma}_{min} = 0.002$, $\bm{\sigma}_{max} = 80$, $\rho = 7$ and $N=32$. Note that we use $\bm{\sigma}$ and $\mathbf{t}$ for EDM to avoid confusion as $\sigma$ and $t$ are also used in our formulations.  To follow the terminology in Algorithm \ref{alg:gfcg0}, $N$ in original EDM is denoted as $T$, sampling steps $i=0,1,\ldots$ is denoted as $t=T,T-1,\ldots$ and noise schedule $\bm{\sigma}(\bm{t}_0), \bm{\sigma}(\bm{t}_1),\ldots$ is denoted as $\sigma_T,\sigma_{T-1},\ldots$ and the noise schedule is calculated as follows:
\begin{eqnarray}\label{eq:sigma}
\sigma_t= \begin{cases}
\left({\bm{\sigma}_{max}}^{\frac{1}{\rho}}+\frac{T-t}{T-1}({\bm{\sigma}_{min}}^{\frac{1}{\rho}}-{\bm{\sigma}_{max}}^{\frac{1}{\rho}})\right)^{\rho} & t > 0 \\
0 & t = 0  
\end{cases}
\end{eqnarray}

Equation \ref{eq:x0} for $\widehat{x}_0$ estimation is replaced by a multi-step denoising process, as also discussed in Section \ref{subsec:ccig}. This modification is detailed in Algorithm \ref{alg:gfcg1}. Although this introduces a few additional NFEs, the impact is minimal since the multi-step estimation is only required once. For instance, in the case of GFCG\textsubscript{ATG}, the parameters in Algorithm \ref{alg:gfcg1} are set as $M_{B} = \text{ATG}$, $\bm{\sigma}_{min}' = 0.002$, $\rho' = 7$ and $N'=4$. As for $s_{cp}$, it is set to the maximum so that only one classifier prediction is used for all mixed and additive GFCG methods included in Table~\ref{tab:edm2}.

\subsection{Text-to-Image Generation: Text Prompts}
\label{sec:p}
For the main quantitative experiments of text-to-image generations using GFCG and other gradient-free guidance methods, a set of generic prompts are used based on the realistic distribution of the Birds Species dataset.  This is designed to minimize the bias between the real and generated images so the \FDD metric could be more reliable in quantitative assessment.  Each of the Set of following 8 generic text prompts was used to generate 10 samples for each bird species and quantitative results are reported in Table \ref{tab:birds}.
\begin{itemize}

\setlength\itemsep{0.05em}
\item[$\bullet$] \textit{a close up photo of a bird, [bird species]}
\item[$\bullet$] \textit{a close up bird photo, [bird species]}
\item[$\bullet$] \textit{a close up picture of a bird, [bird species]}
\item[$\bullet$] \textit{a close up bird picture, [bird species]}
\item[$\bullet$] \textit{a full body photo of a bird, [bird species]}
\item[$\bullet$] \textit{a full body bird photo, [bird species]}
\item[$\bullet$] \textit{a full body picture of a bird, [bird species]}
\item[$\bullet$] \textit{a full body bird picture, [bird species]}

\end{itemize}

A set of detailed text prompts was used to generate samples, 5 per prompt for each species, for the ablation study reported in Figure \ref{fig:sd15_ad}.  These detailed prompts are not realistic for all species, as the roadrunner in Figure \ref{fig:front} doesn't perch on tree branches in real life.  Besides, the descriptions below only cover part of all natural habitats.  As \FDD is not applicable for this test given these biases between two distributions, only Precision scores are reported.
\begin{itemize}

\setlength\itemsep{0.05em}
\item[$\bullet$] \textit{a photo of a bird perching on a tree branch with flowers blooming around it, [bird species]}
\item[$\bullet$] \textit{a close up photo of a flying bird with fish in its claws, [bird species]}
\item[$\bullet$] \textit{a photo of a bird eating red berry when standing on a rock, [bird species]}
\item[$\bullet$] \textit{a photo of a bird walking on the beach on a raining day, [bird species]}
\item[$\bullet$] \textit{a photo of a bird, [bird species], perching on a tree branch with flowers blooming around it}
\item[$\bullet$] \textit{a close up photo of a flying bird, [bird species], with fish in its claws}
\item[$\bullet$] \textit{a photo of a bird, [bird species], eating red berry when standing on a rock}
\item[$\bullet$] \textit{a photo of a bird, [bird species], walking on the beach on a raining day}

\end{itemize}

\subsection{SEG implementation details}
For the implementation of SEG in the EDM2 codebase, we consider $\sigma$ = 100 for the Gaussian Blur. This is applied in the EDM2 UNet's following blocks in the guidance network: 
\begin{verbatim}
8x8_block1
8x8_block2
8x8_in0
8x8_block0
\end{verbatim}
Other values of $\sigma$ (=10 and 1e6) were also considered but the observations are very similar. 

The SEG implementation in SD is similar to \cite{hong_arxiv_2024}.

\section{More Experimental Results}
\subsection{Effects of Random Seed Variation}
\begin{figure}[thb]
\captionsetup[subfigure]{labelformat=empty}
  \centering
  \vspace{-1pt}
  \includegraphics[width=0.45\textwidth, interpolate=false]{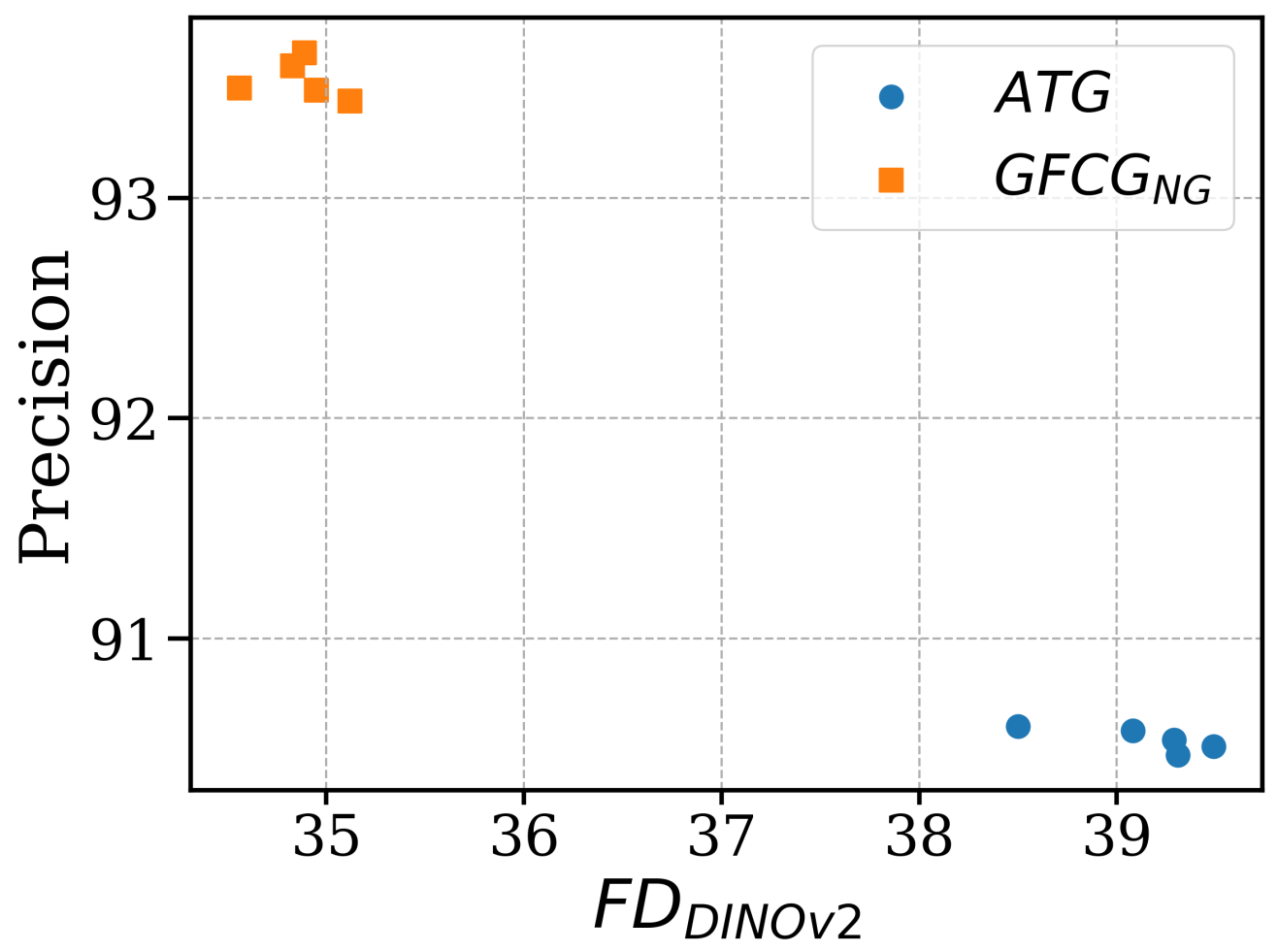} 
  \vspace{-5pt}
  \caption{Ablation study for impact on Precision and \FDD metric against random seed variation.}
  \label{fig:randomseed}
  \vspace{-15pt}
\end{figure}

The results presented in Table \ref{tab:edm2} of the main paper were generated using the same random seed for image generation. As the random seeds used in the ATG study~\cite{karras_arxiv_2024} were not disclosed, we were unable to exactly replicate the reported \FDD metric. To demonstrate that the improvement in the \FDD metric is independent of random seed choice, we have illustrated the variation in \FDD and Precision with random seeds for both ATG and GFCG\textsubscript{NG} methods in Figure \ref{fig:randomseed}. The results show a clear distinction between the two methods in terms of Precision and \FDD metrics, indicating that GFCG\textsubscript{NG} consistently outperforms ATG, regardless of the random seed used.

\subsection{Effects of Guidance Model}
The experiments detailed in the main paper involving the EDM2-S and EDM2-XXL models utilize guidance models (XS, T/16) and (M, T/3.5) respectively for the GFCG and SEG experiments, similar to ATG~\cite{karras_arxiv_2024}. These guidance models, with reduced capacity and training, are readily accessible thanks to the publicly available EDM2 codebase~\cite{karras_cvpr_2024}. However, this availability may not extend to other class-conditional or text-to-image generation diffusion models. Table \ref{tab:edm2gm} presents the \FDD and Precision metrics for the GFCG\textsubscript{NG} method, based on the capacity and training of the guidance model. The hyperparameters for GFCG, $\alpha$, $\beta$, and $t_{s}$, are set to $0.85$, $1.25$, and $17$, respectively. Similar to ATG, reducing training significantly impacts the \FDD metric, while reducing capacity only results in the worst performance. The highest precision is achieved when using the same guidance model as the main model with some degradation in \FDD metric.     
\begin{table}[thb]
	\centering
	\footnotesize
	\setlength{\tabcolsep}{2pt}
\vspace{0pt}
	\caption{Study impact of guidance model capacity and training for GFCG\textsubscript{NG} method using {\FDD} and Precision metrics for 50000 generated samples from EDM2-S, assessed with the ImageNet dataset. The best in each metric is highlighted in \textbf{bold} and the second best is marked with \underline{underline}.}
\vspace{0pt}
	\begin{tabular}{r|cc|ccc} 
		\hline
		{} & {\FDD$\downarrow$} & {Precision$\uparrow$} & {$M_{g}$} & {$EMA_{m}$} & {$EMA_{g}$} \\
		\hline \hline
        {Reduce capacity} & {51.21} & {93.3\%} & {(XS,T)} & {0.085} & {0.085} \\
        {Reduce training} & {\underline{39.36}} & {\underline{93.6\%}} & {(S,T/16)} & {0.085} & {0.170} \\
        {Same both} & {47.79} & {\textbf{94.0\%}} & {(S,T)} & {0.085} & {0.085} \\
        {Reduce both} & {\textbf{36.93}} & {93.3\%} & {(XS,T/16)} & {0.085} & {0.165} \\
		\hline	\end{tabular}
\label{tab:edm2gm}
\vspace{-9pt}
\end{table}

\begin{figure*}[tbh]
  \centering
  \vspace{-1pt}
  %\captionsetup[subfigure]{font=footnotesize, labelformat=empty}
  \begin{subfigure}{0.27\linewidth}
    \includegraphics[height=4.1cm, interpolate=false]{CVPR2025/images/sd15_alpha.png}
    \caption{\hspace{0.1cm} $\alpha$}
  \end{subfigure}
  \hspace{2pt}
  % \hfill
  \begin{subfigure}{0.22\linewidth}
    \includegraphics[height=4.1cm, interpolate=false]{CVPR2025/images/sd15_fcp.png}
    \caption{\hspace{0.1cm} $s_{cp}$}
  \end{subfigure}
  \hspace{2pt}
  % \hfill
  \begin{subfigure}{0.27\linewidth}
    \includegraphics[height=4.1cm, interpolate=false]{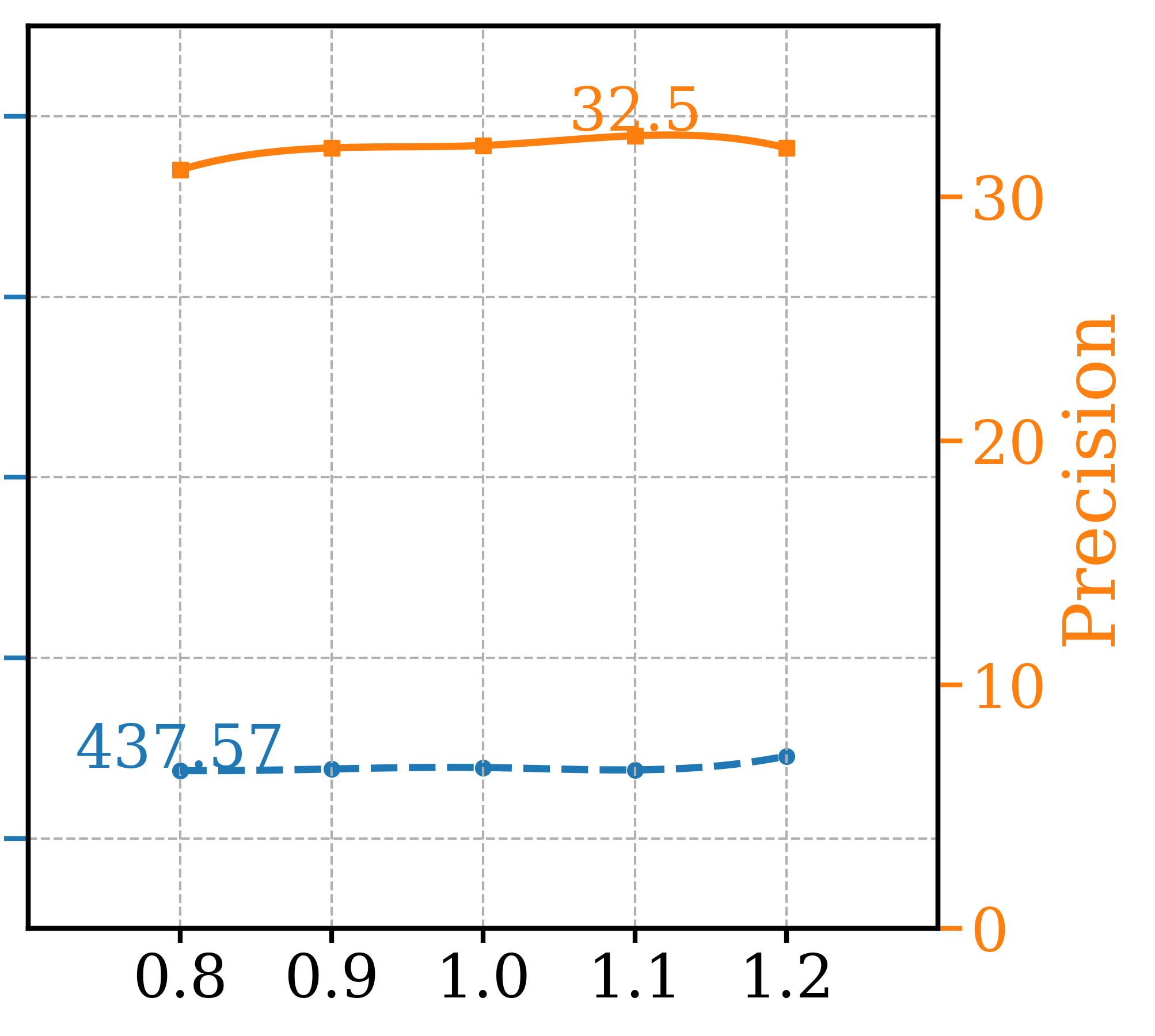}
    \caption{\hspace{0.1cm} $\beta$}
  \end{subfigure}
  \vspace{-5pt}
  \caption{Ablation studies for GFCG method: text-to-image generations (8,400 samples)}
  \label{fig:sd15_gfcg_supp}
  \vspace{0pt}
\end{figure*}

\begin{figure*}[tbh]
  \centering
  \vspace{-1pt}
  %\captionsetup[subfigure]{font=footnotesize, labelformat=empty}
  \begin{subfigure}{0.27\linewidth}
    \includegraphics[height=4.1cm, interpolate=false]{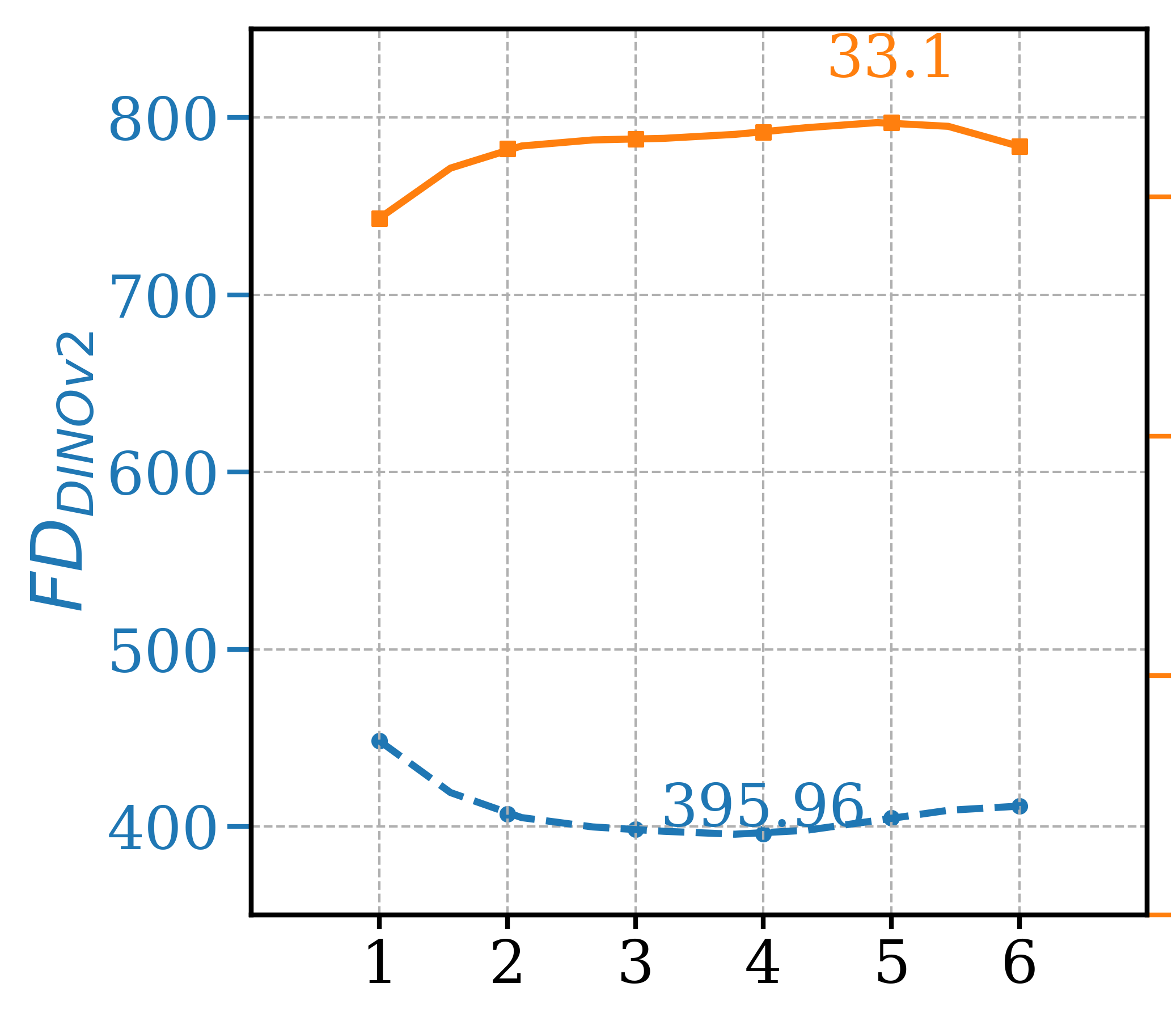}
    \caption{\hspace{0.1cm} $\omega_{CFG}$}
  \end{subfigure}
  \hspace{2pt}
  % \hfill
  \begin{subfigure}{0.27\linewidth}
    \includegraphics[height=4.1cm, interpolate=false]{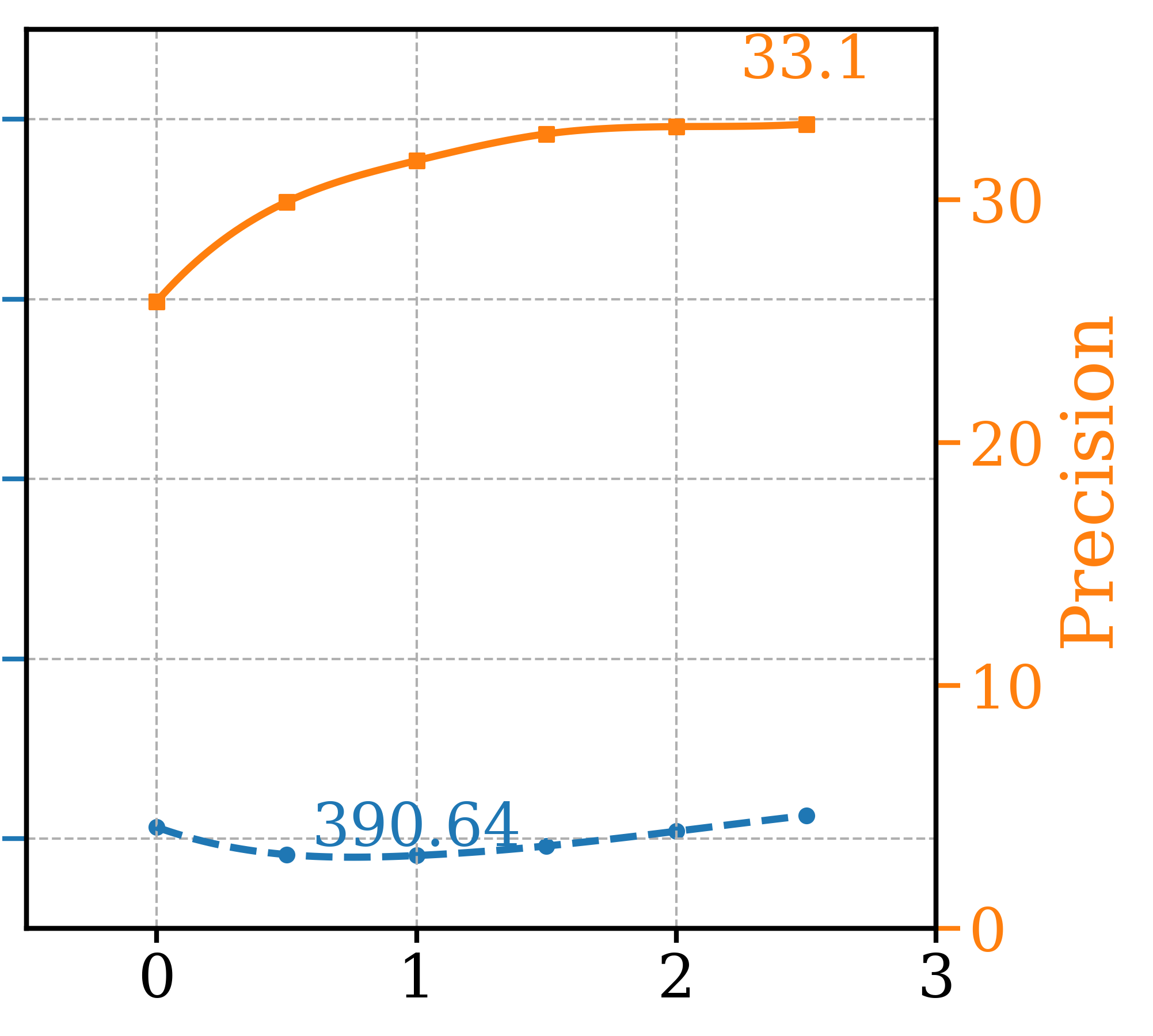}
    \caption{\hspace{0.1cm} $\alpha$}
  \end{subfigure}
  \vspace{-5pt}
  \caption{Ablation studies for GFCG+CFG method: text-to-image generations (8,400 samples)}
  \label{fig:sd15_gfcg_cfg_supp}
  \vspace{0pt}
\end{figure*}

\subsection{Effects of Classifier Model}
Different classifier models, with varying sizes and top-1 and top-5 accuracies on ImageNet-1k (acc@1 and acc@5 in Table \ref{tab:edm2gm}), were considered in the ablation study for classifier predictions in GFCG-based methods. ResNet-18, which is one-fourth the size of ResNet-101 used in the main tests, achieved a high precision of 93\% compared to ATG's 90.6\%, while maintaining a similar \FDD to ATG. ResNet-101 exhibited comparable \FDD and precision metrics to ResNet-152, as shown in Table \ref{tab:edm2gm}, but with a smaller model size, and was thus selected for the main experiments in the paper.

\subsection{Effects of $\widehat{x}_{0}$ Estimation Methods}
For all the mixed and additive GFCG methods presented in Table \ref{tab:edm2}, $s_{cp}$ is set to its maximum value and $T'$ is set to 4 for $\widehat{x}_{0}$ estimation. Although this introduces 7 additional NFEs, which is minimal compared to the 63 NFEs, it results in a significant boost in precision and some improvement in \FDD as well (see Table \ref{tab:edm2}). We explore two methods to further reduce the NFEs. The first method is to reduce the number of steps for $\widehat{x}_{0}$ estimation. The second method is to use a smaller and lower-trained guidance model as the main model for $\widehat{x}_{0}$ estimation. For instance, the guidance model used in the majority of the EDM2-S experiments is EDM2-XS, trained for T/16. If $M_{B} = \text{ATG}$, then line 32 in Algorithm \ref{alg:gfcg1} would change to $D_1 \leftarrow D^{g}_{\phi}(x_t, \sigma_t, c{des})$, $D_2 \leftarrow D^{g}_{\phi}(x_t, \sigma_t, c{des})$ and $\widehat{D} \leftarrow \omega_{ATG} D_1 - (\omega_{ATG}-1) D_2$, which essentially applies the NG method using the guidance model only for $\widehat{x}_{0}$ estimation. As a smaller model is used for $\widehat{x}_{0}$ estimation, we ignore the NFEs added by this method. The results of these two methods compared to the main paper results are presented in Table \ref{tab:edm2x0}.

\begin{table}[h]
	\centering
	\footnotesize
	\setlength{\tabcolsep}{2pt}
\vspace{0pt}
	\caption{Study impact of choice of classifier model for GFCG\textsubscript{NG} method using {\FDD} and Precision metrics for 50000 generated samples from EDM2-S, assessed with the ImageNet dataset. The best in each metric is highlighted in \textbf{bold} and the second best is marked with \underline{underline}.}
\vspace{0pt}
	\begin{tabular}{r|cc|cccc} 
		\hline
		{Classifier} & {\FDD$\downarrow$} & {Precision$\uparrow$} & {Mparams} & {Gflops} & {acc@1} & {acc@5} \\
		\hline \hline
        {ResNet-18} & {39.54} & {93.0\%} & {11.7} & {1.81} & {69.8\%} & {89.1\%} \\
        {ResNet-34} & {40.70} & {93.0\%} & {21.8} & {3.66} & {73.3\%} & {91.4\%} \\
        {ResNet-50} & {37.08} & {93.2\%} & {25.6} & {4.09} & {80.9\%} & {95.4\%} \\
        {ResNet-101} & {\underline{36.93}} & {\underline{93.3\%}} & {44.5} & {7.80} & {81.9\%} & {95.8\%} \\
        {ResNet-152} & {\textbf{36.90}} & {\textbf{93.3\%}} & {60.2} & {11.51} & {82.3\%} & {96.0\%} \\
		\hline	\end{tabular}
\label{tab:edm2gm}
\vspace{-9pt}
\end{table}

\begin{figure*}[tbh]
  \centering
  \vspace{-1pt}
  %\captionsetup[subfigure]{font=footnotesize, labelformat=empty}
  \begin{subfigure}{0.27\linewidth}
    \includegraphics[height=4.1cm, interpolate=false]{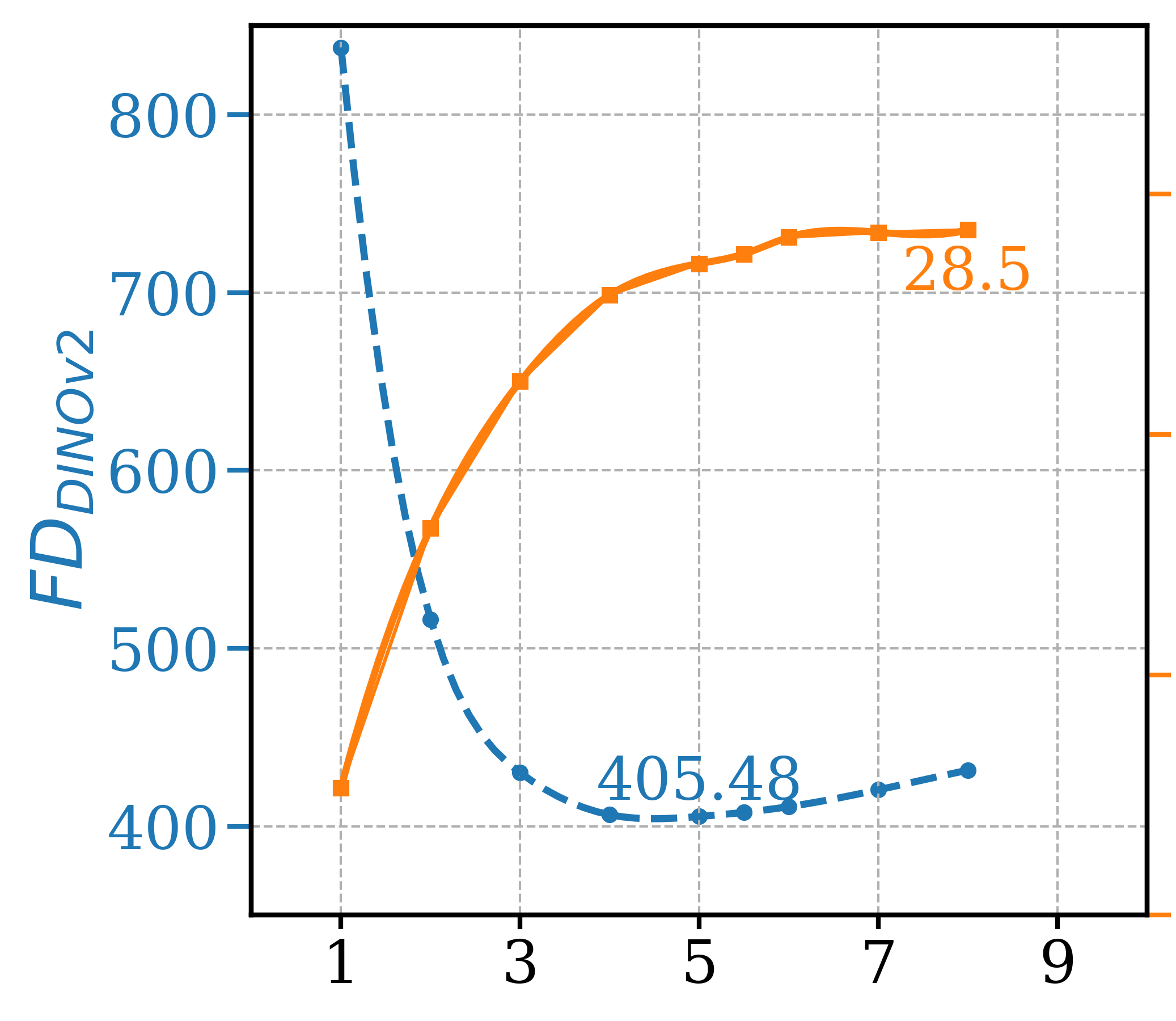}
    \caption{\hspace{0.1cm} $\omega_{CFG}$}
  \end{subfigure}
  \hspace{2pt}
  % \hfill
  \begin{subfigure}{0.22\linewidth}
    \includegraphics[height=4.1cm, interpolate=false]{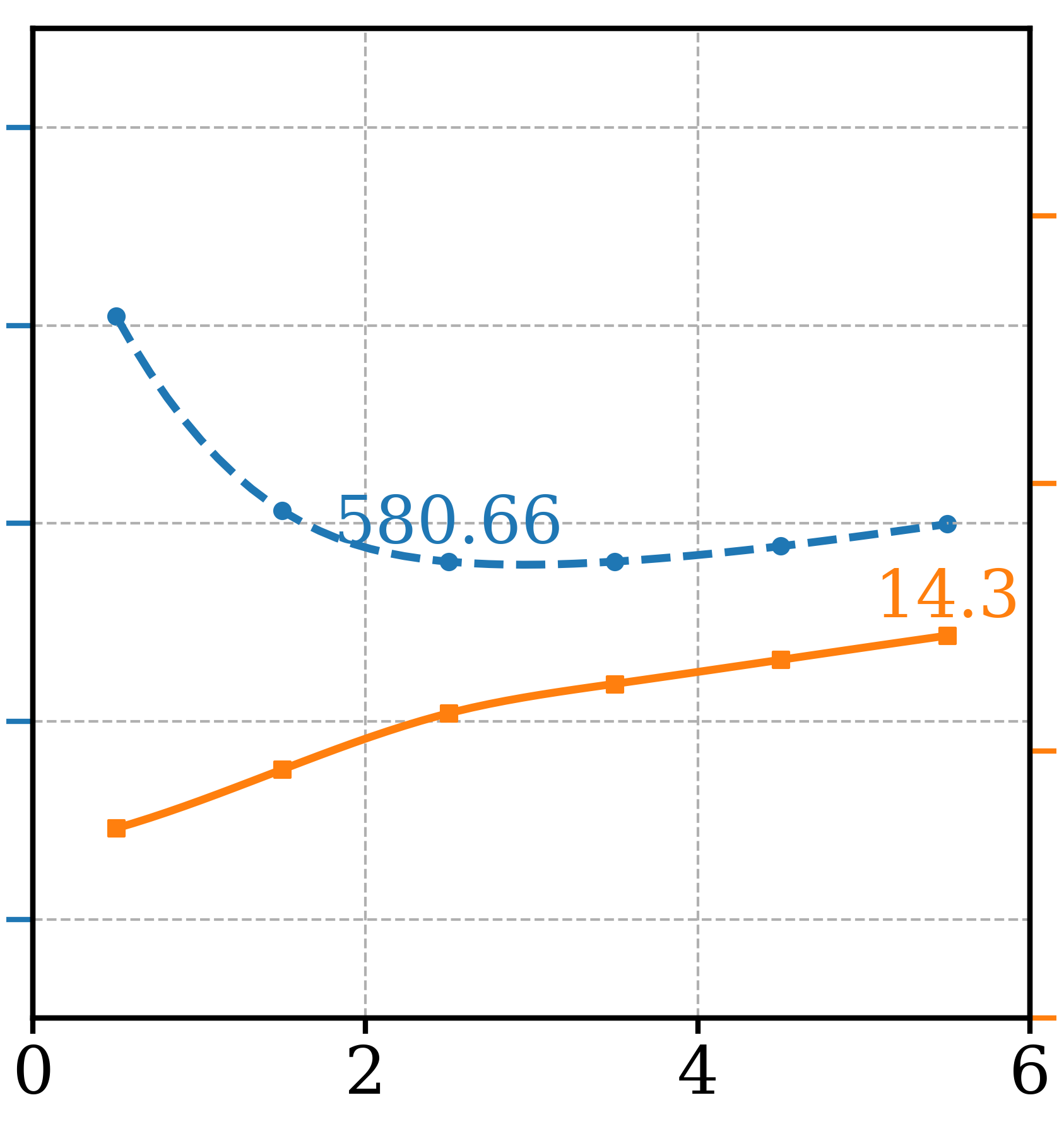}
    \caption{\hspace{0.1cm} $\omega_{PAG}$}
  \end{subfigure}
  \hspace{2pt}
  % \hfill
  \begin{subfigure}{0.27\linewidth}
    \includegraphics[height=4.1cm, interpolate=false]{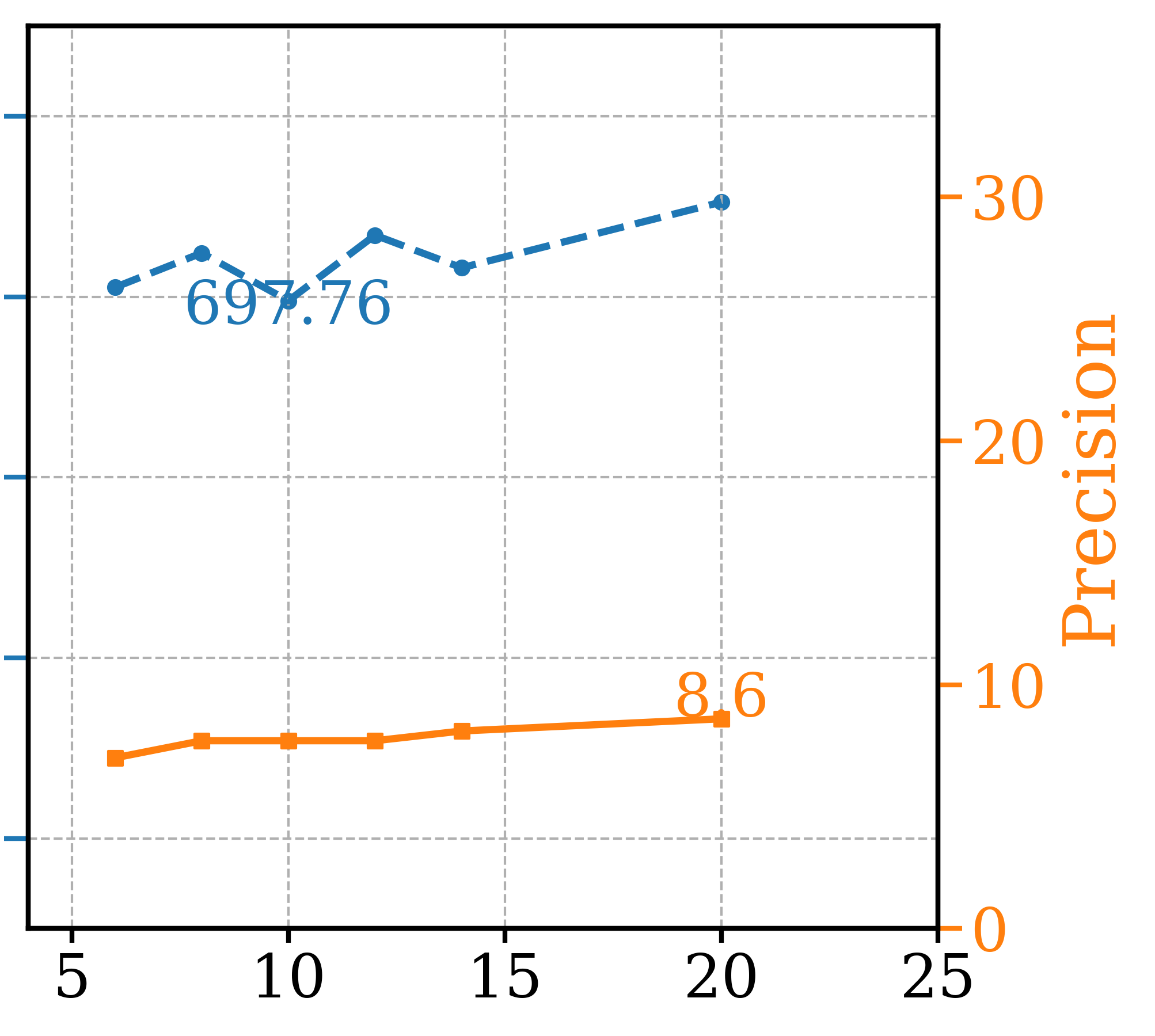}
    \caption{\hspace{0.1cm} $\omega_{SEG}$}
  \end{subfigure}
  \vspace{-5pt}
  \caption{Ablation studies for other guidance methods reported in Table \ref{tab:birds} for text-to-image generations}
  \label{fig:sd15_others_supp}
  \vspace{0pt}
\end{figure*}

\begin{table}[h]
	\centering
	\footnotesize
	\setlength{\tabcolsep}{3pt}
\vspace{0pt}
	\caption{Study impact of methods to reduce NFEs for $\widehat{x}_{0}$ estimation for GFCG\textsubscript{NG} method using {\FDD} and Precision metrics for 50000 generated samples from EDM2-S, assessed with the ImageNet dataset. The best in each metric is highlighted in \textbf{bold} and the second best is marked with \underline{underline}.}
\vspace{0pt}
	\begin{tabular}{r|cc|ccc} 
		\hline
		{} & {\FDD$\downarrow$} & {Precision$\uparrow$} & {$\bm{\sigma}_{min}'$} & {$T'$} & {NFEs} \vspace{1pt} \\
		\hline \hline
        {Method 1} & {39.11} & {92.7\%} & {-} & {1} & {64} \\
        {} & {37.84} & {\underline{93.1\%}} & {1.0} & {2} & {66} \\
        \hline
        {Method 2} & {\textbf{36.05}} & {92.5\%} & {0.002} & {4} & {63} \\
        \hline
        {Main Paper} & {\underline{36.93}} & {\textbf{93.3\%}} & {0.002} & {4} & {70} \\
		\hline	\end{tabular}
\label{tab:edm2x0}
\vspace{-9pt}
\end{table}

\subsection{Stochastic Reference Class Sampling}
As explained in Equation \ref{eq:pref}, a stochastic reference class can be sampled each time a classifier prediction is applied.  It improves sample quality when there are frequent classifier predictions, i.e., $s_{cp}$ is small.  Based on that, the experimental results of text-to-image generations in Table ~\ref{tab:birds} are conducted with this enabled.  For comparison, we compare GFCG methods with stochastic reference class sampling to their counterparts with determinist reference class and the quantitative results are included in Table \ref{tab:srcs}.   It shows that the stochastic methods are better than their deterministic counterparts in overall performance considering both \FDD and Precision.

\begin{table}[h]
	\centering
	\footnotesize
	\setlength{\tabcolsep}{3pt}
\vspace{0pt}
	\caption{Study impact of stochastic reference class sampling comparing to deterministic reference class in text-to-image generations.}
\vspace{0pt}
	\begin{tabular}{rr|cc} 
		\hline
		{Method} & {$c_{ref}$} & {\FDD$\downarrow$} & {Precision$\uparrow$} \\
		\hline \hline
        {GFCG} & {Stochastic} & {418.8} & {32.3\%} \\
        {} & {Deterministic} & {428.8} & {32.4\%} \\
        \hline
        {GFCG\textsubscript{CFG}} & {Stochastic} & {392.3} & {30.2\%} \\
        {} & {Deterministic} & {394.7} & {29.1\%} \\
        \hline
        {GFCG+CFG} & {Stochastic} & {379.2} & {32.4\%} \\
        {} & {Deterministic} & {377.6} & {31.6\%} \\
		\hline	\end{tabular}
\label{tab:srcs}
\vspace{-9pt}
\end{table}

\subsection{Additional Ablation Studies: Text-to-Image}

A full range of ablation studies were conducted for text-to-image experiments using SD, including GFCG and other guidance methods.  Three main ones are included in the main paper. All three key settings of GFCG,  $\alpha$, and $s_{cp}$ included in Figure \ref{fig:sd15_a}, and the additional $\beta$, are plotted together in Figure \ref{fig:sd15_gfcg_supp} for full comparison. As evident, $\alpha$ has the highest impact and $\beta$ the least.  For $s_{cp}$, the Precision value is the highest when it is set as the minimum of 1, while achieving the lowest \FDD too.

For the mixed GFCG$_{CFG}$ method, the only key variable is $t_s$ as $\omega_{CFG}$ for CFG and $\alpha$ for GFCG are using the same optimal setting of each respectively.  The results for $t_s$ is already included in Figure ~\ref{fig:sd15_a}.  For the additive method of GFCG+CFG, $\omega_{CFG}$ and $\alpha$ are investigated to find the optimal settings.  As shown in Figure \ref{fig:sd15_gfcg_cfg_supp}, the optimal values are around 4.0 and 1.5, lower than the settings of 5.5 and 2.0 when optimized for CFG and GFCG individually.
%we undertake ablation studies to investigate the effect of $\omega$, $t_s$ and $\alpha$ and present the results in Figure \ref{fig:sd15_gfcg_cfg_supp}. Comparing the three figures \ref{fig:sd15_gfcg_supp}, \ref{fig:sd15_others_supp} and \ref{fig:sd15_gfcg_cfg_supp}, we note that when GFCG is used, the \FDD is the lowest, with no significant compromise in Precision. This demonstrates the robustness of GFCG for text-to-image generation models. 

For fair comparison, we also study the effect of the guidance scale $\omega$ for CFG, PAG and SEG in Figure \ref{fig:sd15_others_supp}. For CFG, $\omega$ has a significant role in terms of the \FDD and Precision of the generated images. For PAG, $\omega$ has a lesser influence, and for SEG the impact is the least. Among these three guidance methods, CFG is much superior than PAG and SEG in terms of \FDD and Precision. 

\section{Class-Conditional Visual Examples}
\begin{figure*}[thb]
\captionsetup[subfigure]{labelformat=empty}
  \centering
  \vspace{-1pt}
  \includegraphics[width=0.8\textwidth, interpolate=false]{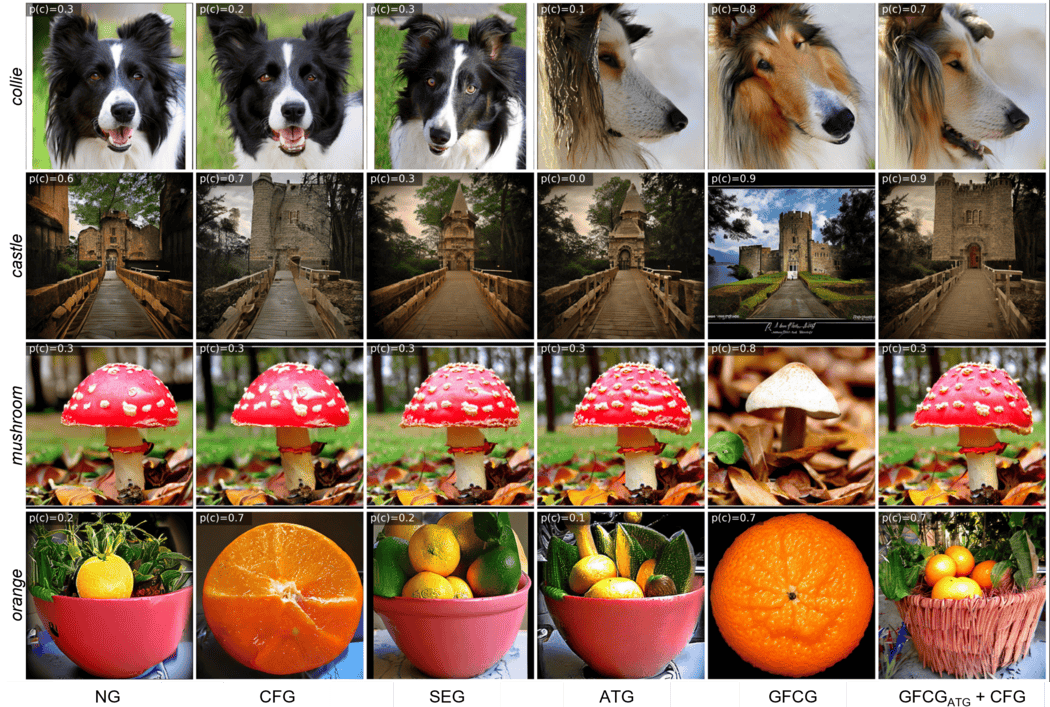} 
  \vspace{-5pt}
  \caption{Visual examples of generated ImageNet class images, comparing GFCG with other guidance methods in EDM2-S sampling. The last column displays examples of the additive method GFCG\textsubscript{ATG}+CFG. While GFCG enhances class accuracy, it sacrifices some diversity. GFCG\textsubscript{ATG}+CFG tries to balance both accuracy and diversity. }
  \label{fig:edm2_supp1}
  \vspace{0pt}
\end{figure*}

\begin{figure*}[thb]
\captionsetup[subfigure]{labelformat=empty}
  \centering
  \vspace{-1pt}
  \includegraphics[width=0.8\textwidth, interpolate=false]{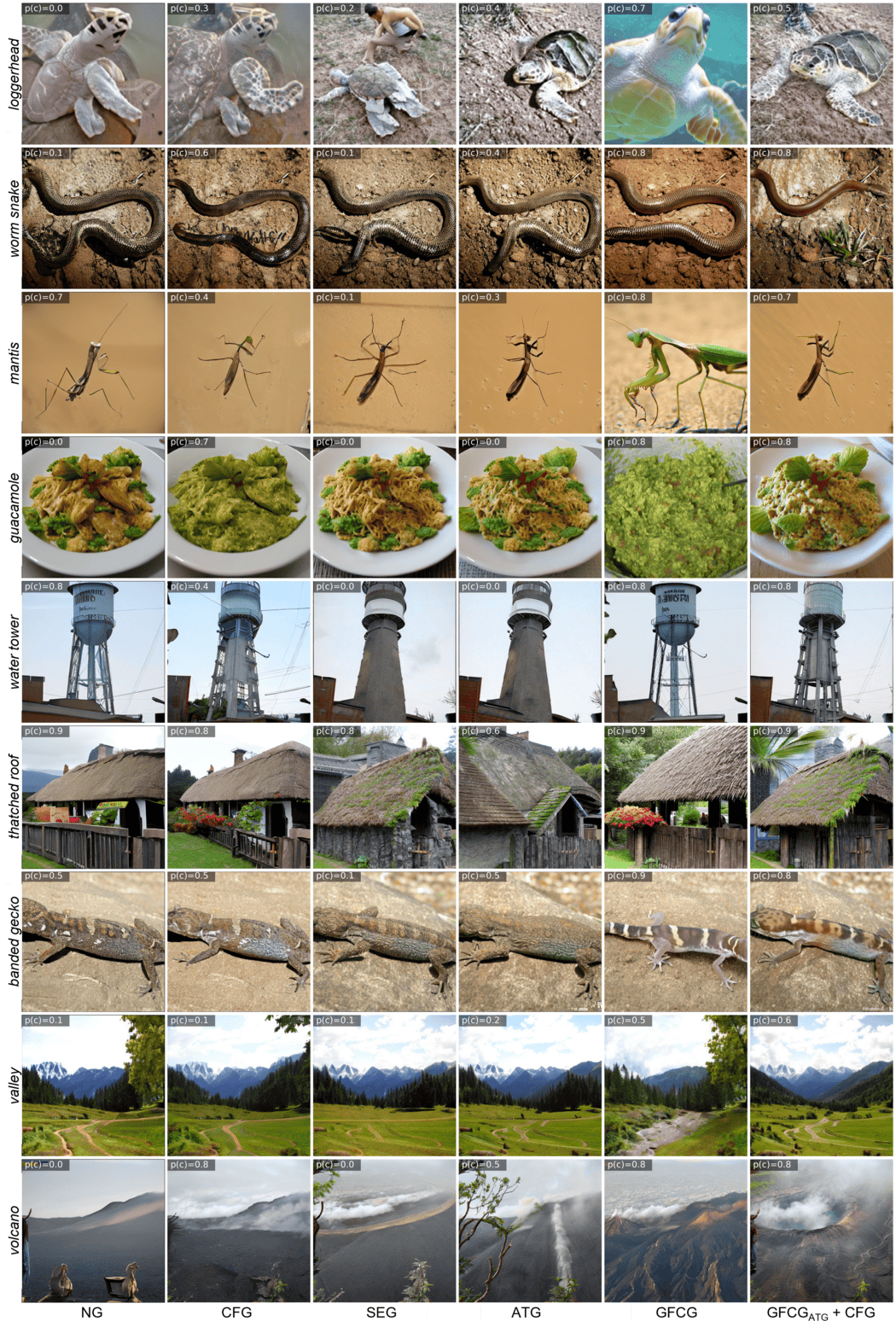} 
  \vspace{-5pt}
  \caption{More visual examples of generated ImageNet class images for different guidance methods in EDM2-S sampling.}
  \label{fig:edm2_supp2}
  \vspace{0pt}
\end{figure*}

Visual examples from class-conditional image generation using existing guidance methods (refer to Table \ref{tab:edm2}) are compared with our GFCG method in EDM2-S sampling, as illustrated in Figure \ref{fig:edm2_supp1}. Additionally, we compare GFCG to the additive method GFCG\textsubscript{ATG}+CFG, which achieves state-of-the-art performance in \FDD for EDM2-S. Further visual examples can be found in Figure \ref{fig:edm2_supp2}. The visual results corroborate the quantitative metrics for Precision, with GFCG-generated images demonstrating a strong alignment with class labels in most cases. This is evident in the examples of \textit{mushroom} and \textit{collie} in Figure \ref{fig:edm2_supp1}, where other guidance methods often confuse \textit{mushroom} with \textit{agaric} and \textit{collie} with \textit{border collie}. However, this precise alignment does result in a slight trade-off in diversity, as seen in the \textit{orange} class example, where GFCG generates a zoomed-in image of a single orange. The additive method GFCG\textsubscript{ATG}+CFG mitigates this issue by balancing diversity and class accuracy, as illustrated in the \textit{orange} class example in Figure \ref{fig:edm2_supp1}, where it generates a bunch of oranges in a basket.

\begin{figure*}[thb]
\captionsetup[subfigure]{labelformat=empty}
  \centering
  \vspace{-1pt}
  \includegraphics[width=0.8\textwidth, interpolate=false]{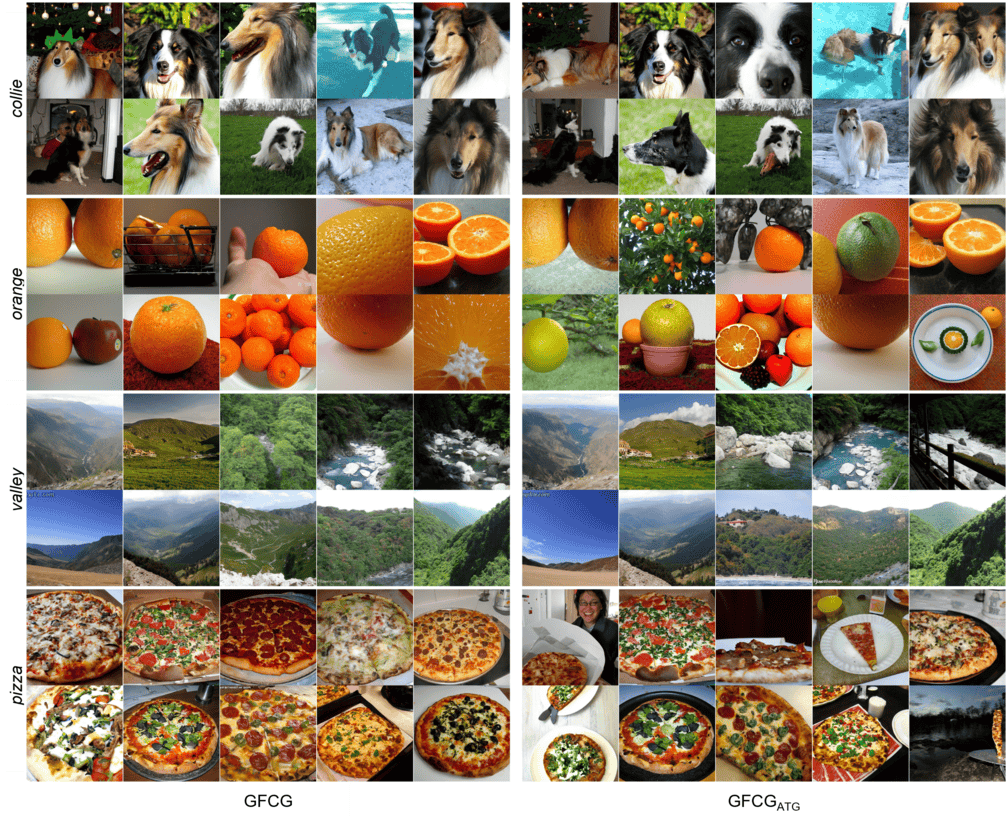} 
  \vspace{-5pt}
  \caption{Visual examples of generated ImageNet class images comparing GFCG and GFCG\textsubscript{ATG} in diversity for EDM2-XXL sampling.}
  \label{fig:edm2_diversity_supp1}
  \vspace{-15pt}
\end{figure*}

\begin{figure*}[thb]
\captionsetup[subfigure]{labelformat=empty}
  \centering
  \vspace{-1pt}
  \includegraphics[width=0.8\textwidth, interpolate=false]{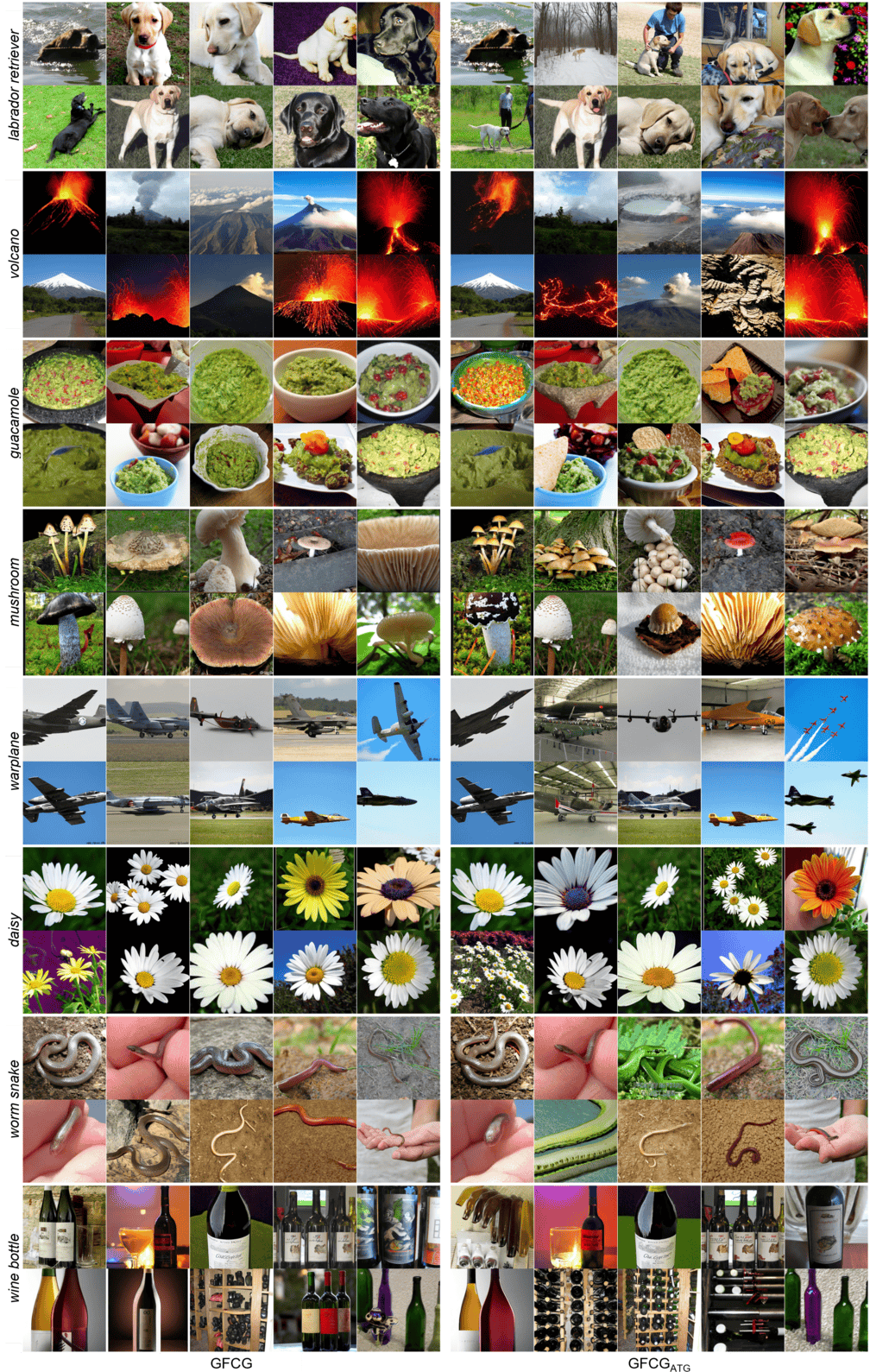} 
  \vspace{-5pt}
  \caption{More visual examples of generated ImageNet class images comparing GFCG and GFCG\textsubscript{ATG} in diversity for EDM2-XXL sampling.}
  \label{fig:edm2_diversity_supp2}
  \vspace{-15pt}
\end{figure*}

To further explore the differences in diversity between GFCG and mixed methods, we compare GFCG with GFCG\textsubscript{ATG}, which achieves state-of-the-art \FDD for EDM2-XXL. We present visual examples displaying 10 images per class for selected classes in Figure \ref{fig:edm2_diversity_supp1}, with additional examples in Figure \ref{fig:edm2_diversity_supp2}. Figure \ref{fig:edm2_diversity_supp1} shows that while GFCG-generated samples align closely with class labels (notably in the \textit{collie} and \textit{orange} cases, which other methods confuse with \textit{border collie} and \textit{lemon}), there is a modest reduction in diversity. For instance, in the \textit{orange} class, GFCG tends to zoom in on the oranges or exclude other fruits, compared to the mixed method. GFCG may also remove or modify background objects which may cause confusion with the target class, as seen in the \textit{pizza} and \textit{valley} classes in Figure \ref{fig:edm2_diversity_supp1}. Mixed methods, particularly with ATG, help preserve diversity while enhancing class accuracy.

\section{Text-to-Image Visual Examples}

For text-to-image generations, the results presented in the main paper were all based on samples from SD 1.5 and more visual examples are included here.  Additionally, we also conducted experiments using another popular model, DeepFloyd IF model\footnote{https://github.com/deep-floyd/IF} from Stability AI.  Some examples are included in Section \ref{sec:if}.

\begin{figure*}[t!]
\captionsetup[subfigure]{font=footnotesize, labelformat=empty}
 \begin{center}
  \begin{subfigure}[b]{0.98\textwidth}
    \centering
      \vspace{-0.1em}
      \includegraphics[width=\textwidth, interpolate=false]{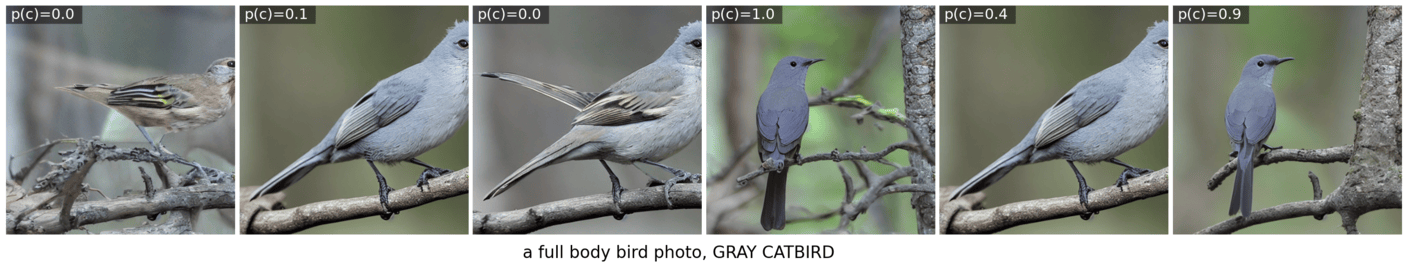}
  \end{subfigure}
  \begin{subfigure}[b]{0.98\textwidth}
    \centering
      \vspace{-0.1em}
      \includegraphics[width=\textwidth, interpolate=false]{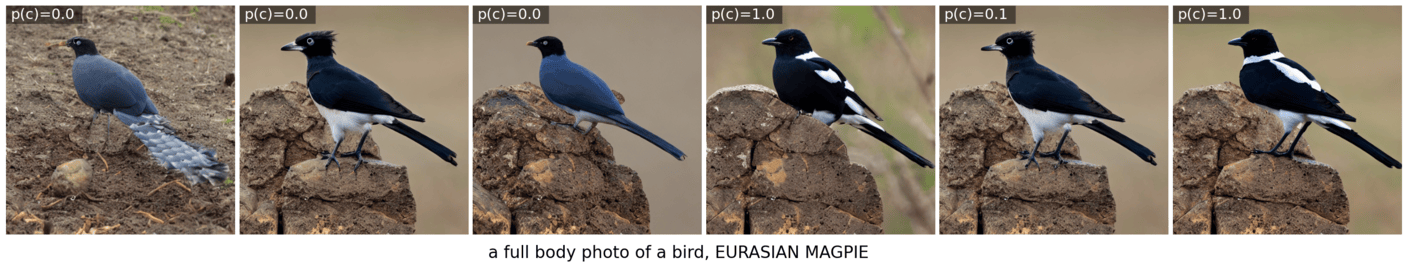}
  \end{subfigure}
  \begin{subfigure}[b]{0.98\textwidth}
    \centering
      \vspace{-0.1em}
      \includegraphics[width=\textwidth, interpolate=false]{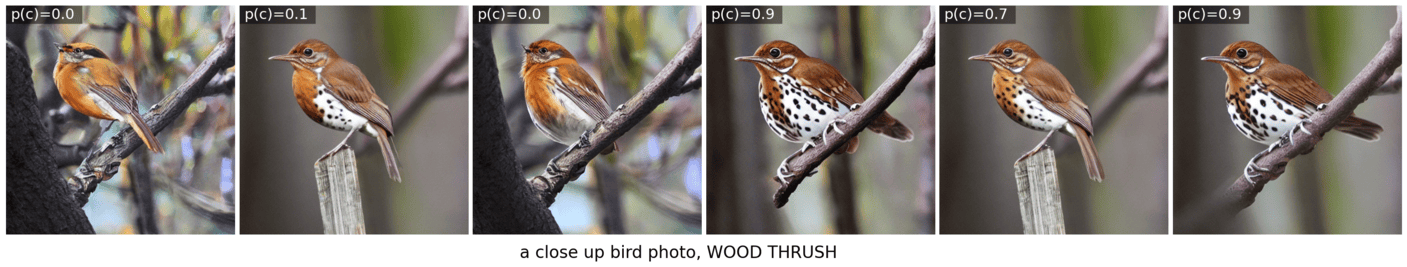}
      \caption{\hspace{1em} NG \hspace{8em} CFG \hspace{8em} PAG \hspace{7em} GFCG \hspace{6em} GFCG\textsubscript{CFG} \hspace{5em} GFCG+CFG}
  \end{subfigure}
 \end{center}
 \vspace{-15pt}
 \caption{Representative visual examples which demonstrate the benefits of GFCG over others in text-to-image generation using generic text prompts: 1) GFCG and GFCG+CFG improve compositional quality to improve class accuracy; 2) They add the right feather color pattern for higher class probability; 3) GFCG\textsubscript{CFG} makes minor improvement in chest spot pattern resulting in higher probability.}
 \label{fig:gtpg}
\end{figure*}

\begin{figure*}[t!]
\captionsetup[subfigure]{font=footnotesize, labelformat=empty}
 \begin{center}
  \begin{subfigure}[b]{0.98\textwidth}
    \centering
      \vspace{-0.1em}
      \includegraphics[width=\textwidth, interpolate=false]{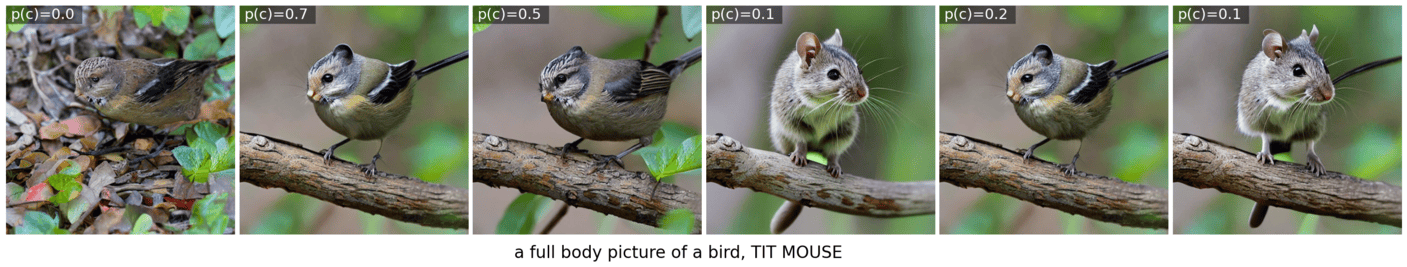}
  \end{subfigure}
  \begin{subfigure}[b]{0.98\textwidth}
    \centering
      \vspace{-0.1em}
      \includegraphics[width=\textwidth, interpolate=false]{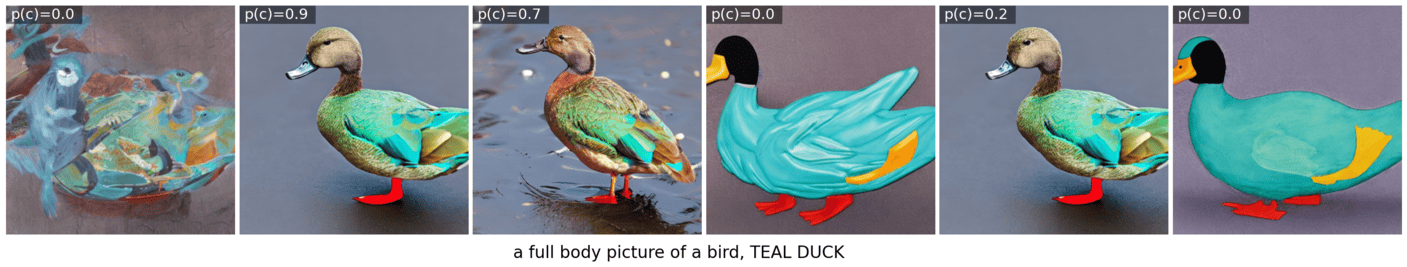}
      \caption{\hspace{1em} NG \hspace{8em} CFG \hspace{8em} PAG \hspace{7em} GFCG \hspace{6em} GFCG\textsubscript{CFG} \hspace{5em} GFCG+CFG}
  \end{subfigure}
 \end{center}
 \vspace{-15pt}
 \caption{Representative visual examples where GFCG fails to improve class accuracy using generic prompts, where the incorrect semantic understanding of \textit{MOUSE} and \textit{TEAL} gets enhanced due to GFCG.}
 \label{fig:gtpb}
\end{figure*}

\subsection{Generic Text Prompts}
Some visual examples from text-to-images generation using the set of generic prompts are included in Figure ~\ref{fig:gtpg}.  The probability of classifying each generated sample as the target class is also included for reference.  In general, the GFCG results have the best class accuracy.  The gained accuracy could be caused by compositional change in the first example, as well as correct anatomic features like feather color in the second.
For GFCG\textsubscript{CFG}, it often maintains the overall composition of CFG and is possible to improve class accuracy even when the changes are negligible with untrained eyes like the first example.  In the case of the last example, minor changes like chest patterns in GFCG\textsubscript{CFG} result in significantly increased probability too.

As shown in Figure ~\ref{fig:gtpb}, GFCG results may end up worse than other methods too.  For the first example of \textit{TIT MOUSE}, the text-to-image model obviously has misunderstood \textit{MOUSE} without recognizing its context as a bird specie.  In the case of CFG, as it is guided away from an unconditional model, it will enhance the wrong features associated with \text{MOUSE}, as well as correct features associate with other keywords like \textit{bird}.  For GFCG, as the enhancement is in the reference to a photo of another bird species, the difference will be focused between \textit{TIT MOUSE} and another bird species which results in more prominent mouse features.  Similar failure happens in the second example where color features related to \textit{TEAL} is magnified.

\begin{figure*}[t!]
\captionsetup[subfigure]{font=footnotesize, labelformat=empty}
 \begin{center}
  \begin{subfigure}[b]{0.98\textwidth}
    \centering
      \vspace{-0.1em}
      \includegraphics[width=\textwidth, interpolate=false]{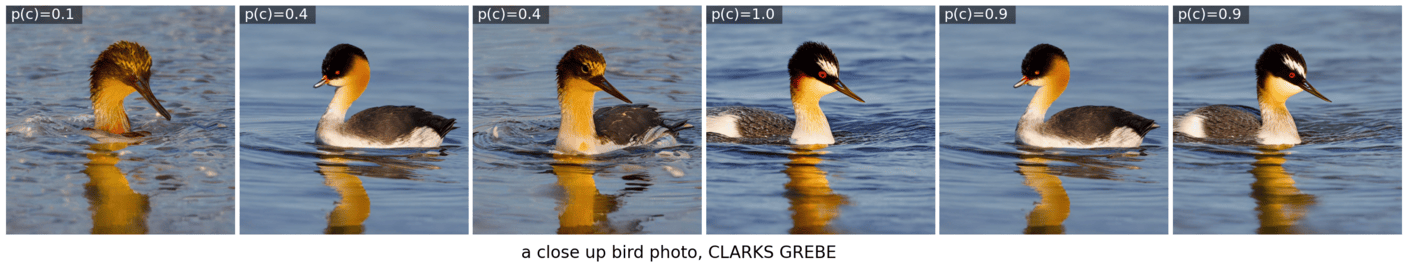}
  \end{subfigure}
  \begin{subfigure}[b]{0.98\textwidth}
    \centering
      \vspace{-0.1em}
      \includegraphics[width=\textwidth, interpolate=false]{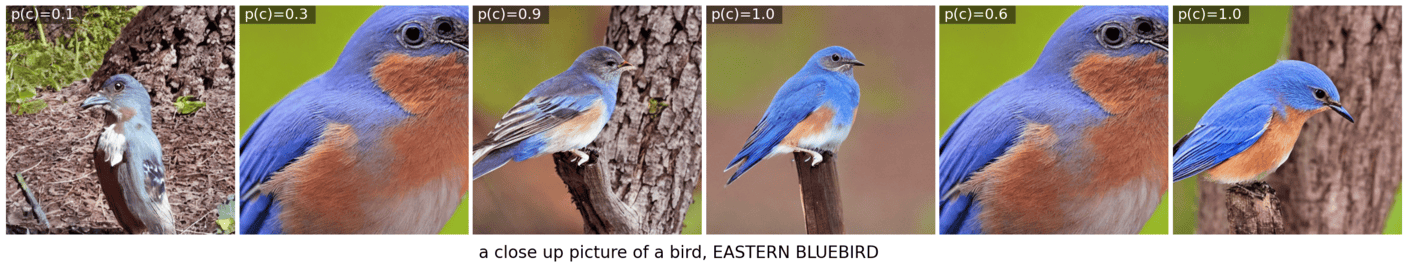}
  \end{subfigure}
  \begin{subfigure}[b]{0.98\textwidth}
    \centering
      \vspace{-0.1em}
      \includegraphics[width=\textwidth, interpolate=false]{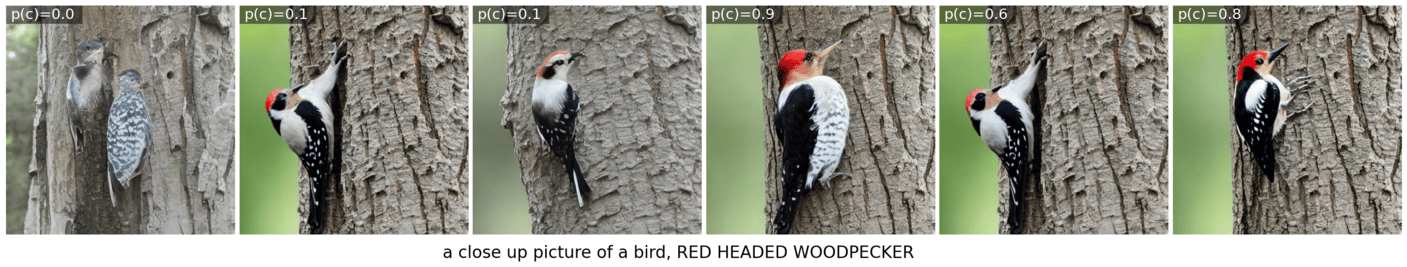}
  \end{subfigure}
  \begin{subfigure}[b]{0.98\textwidth}
    \centering
      \vspace{-0.1em}
      \includegraphics[width=\textwidth, interpolate=false]{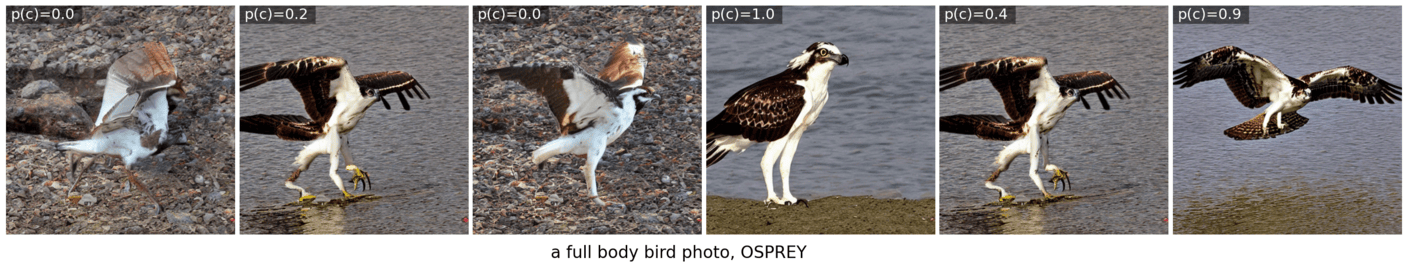}
  \end{subfigure}
  \begin{subfigure}[b]{0.98\textwidth}
    \centering
      \vspace{-0.1em}
      \includegraphics[width=\textwidth, interpolate=false]{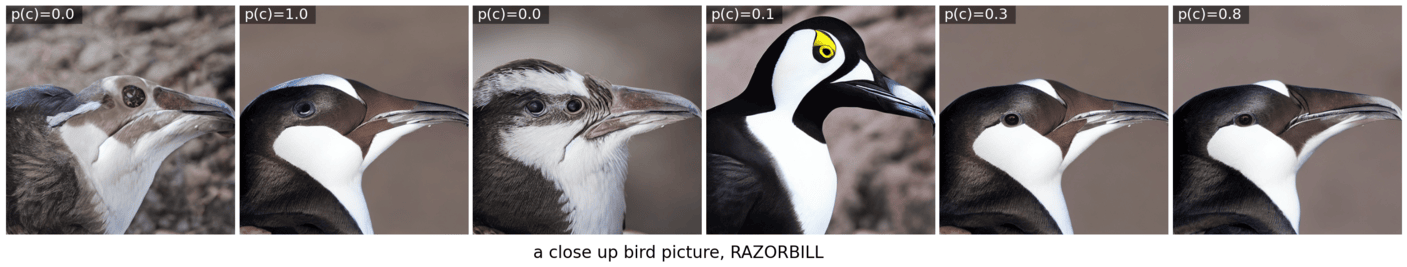}
  \end{subfigure}
  \begin{subfigure}[b]{0.98\textwidth}
    \centering
      \vspace{-0.1em}
      \includegraphics[width=\textwidth, interpolate=false]{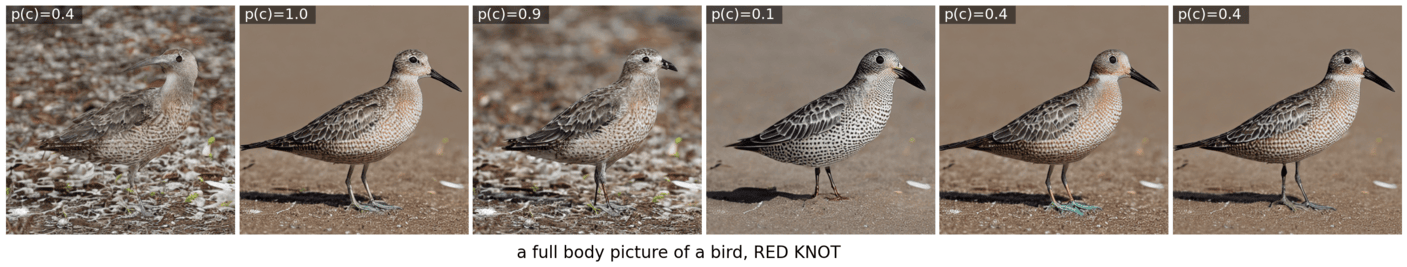}
      \caption{\hspace{1em} NG \hspace{8em} CFG \hspace{8em} PAG \hspace{7em} GFCG \hspace{6em} GFCG\textsubscript{CFG} \hspace{5em} GFCG+CFG}
  \end{subfigure}
 \end{center}
 \vspace{-15pt}
 \caption{More visual examples from SD 1.5 model using generic text prompts.}
 \label{fig:gtpm}
\end{figure*}

More visual examples using generic text prompts are shown in Figure \ref{fig:gtpm} at the end.

\begin{figure*}[t!]
\captionsetup[subfigure]{font=footnotesize, labelformat=empty}
 \begin{center}
  \begin{subfigure}[b]{0.8\textwidth}
    \centering
      \vspace{-0.1em}
      \includegraphics[width=\textwidth, interpolate=false]{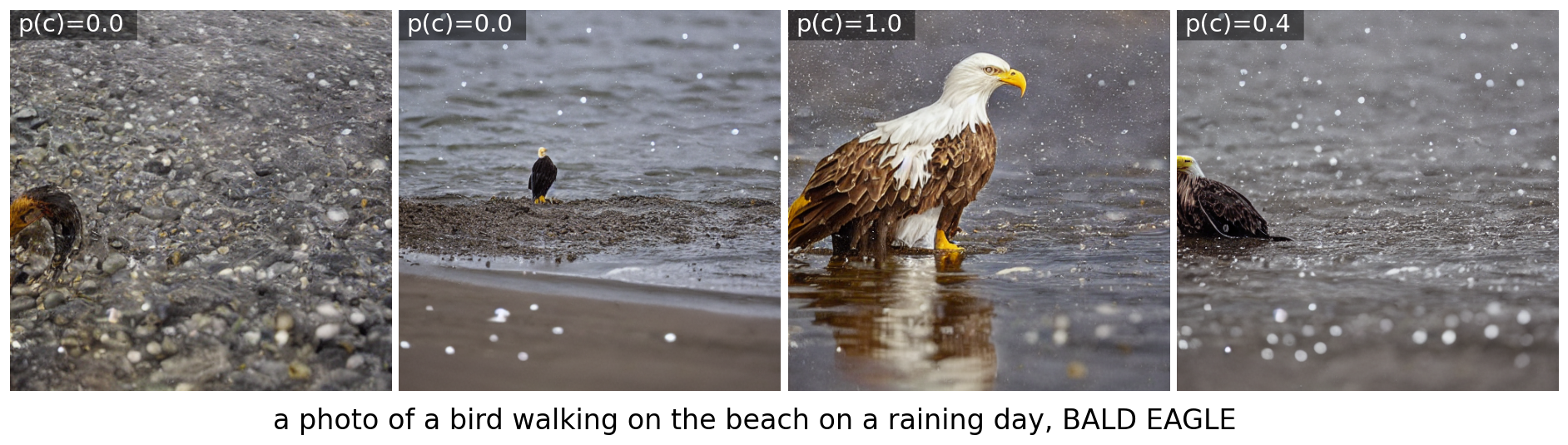}
  \end{subfigure}
  \begin{subfigure}[b]{0.8\textwidth}
    \centering
      \vspace{-0.1em}
      \includegraphics[width=\textwidth, interpolate=false]{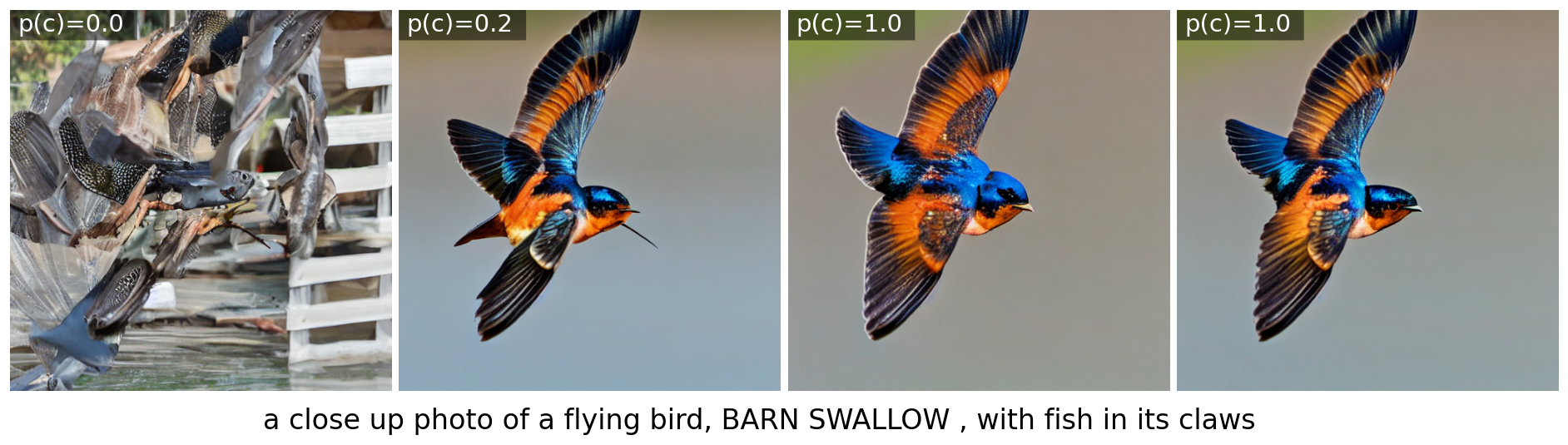}
  \end{subfigure}
  \begin{subfigure}[b]{0.8\textwidth}
    \centering
      \vspace{-0.1em}
      \includegraphics[width=\textwidth, interpolate=false]{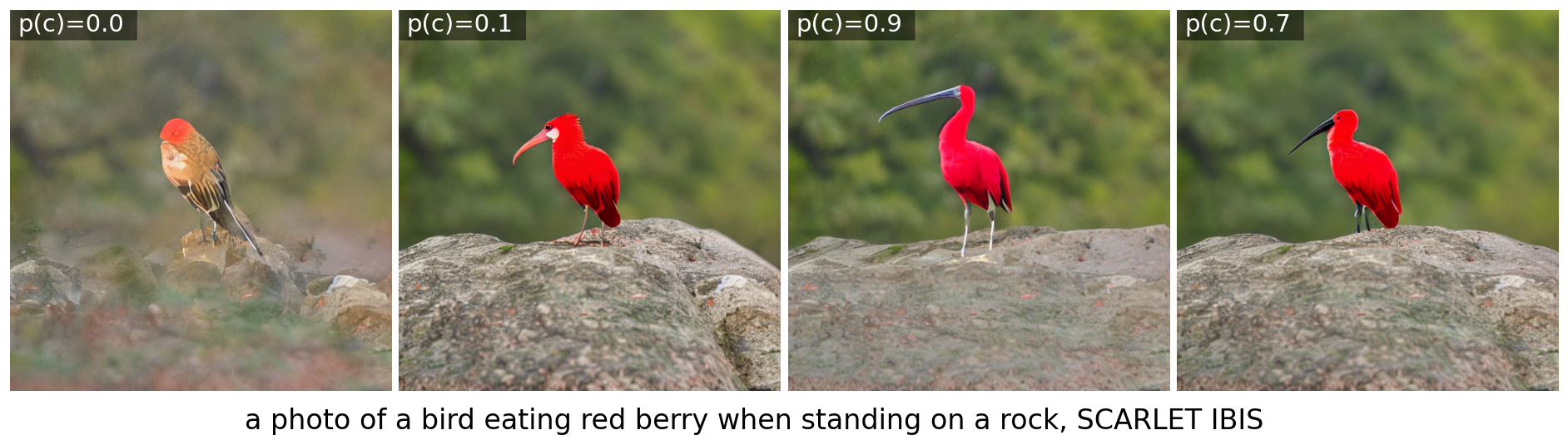}
      \caption{\hspace{1em} NG \hspace{10em} CFG \hspace{9em} GFCG  \hspace{8em} GFCG+CFG}
  \end{subfigure}
 \end{center}
 \vspace{-15pt}
 \caption{Representative visual examples where GFCG improves class accuracy using detailed text prompts: 1) GFCG enhances the bird features while preserving contextual coherence; 2) GFCG and GFCG+CFG correct the flying posture; 3) or render the right beak color.}
 \label{fig:dtpg}
\end{figure*}

\begin{figure*}[t!]
\captionsetup[subfigure]{font=footnotesize, labelformat=empty}
 \begin{center}
  \begin{subfigure}[b]{0.8\textwidth}
    \centering
      \vspace{-0.1em}
      \includegraphics[width=\textwidth, interpolate=false]{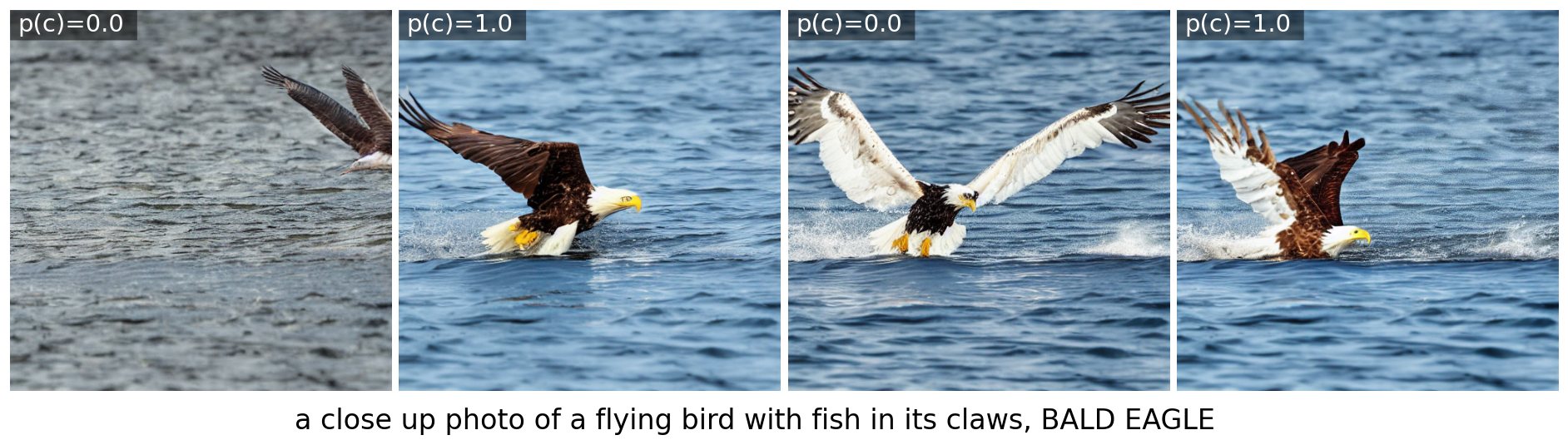}
  \end{subfigure}
  \begin{subfigure}[b]{0.8\textwidth}
    \centering
      \vspace{-0.1em}
      \includegraphics[width=\textwidth, interpolate=false]{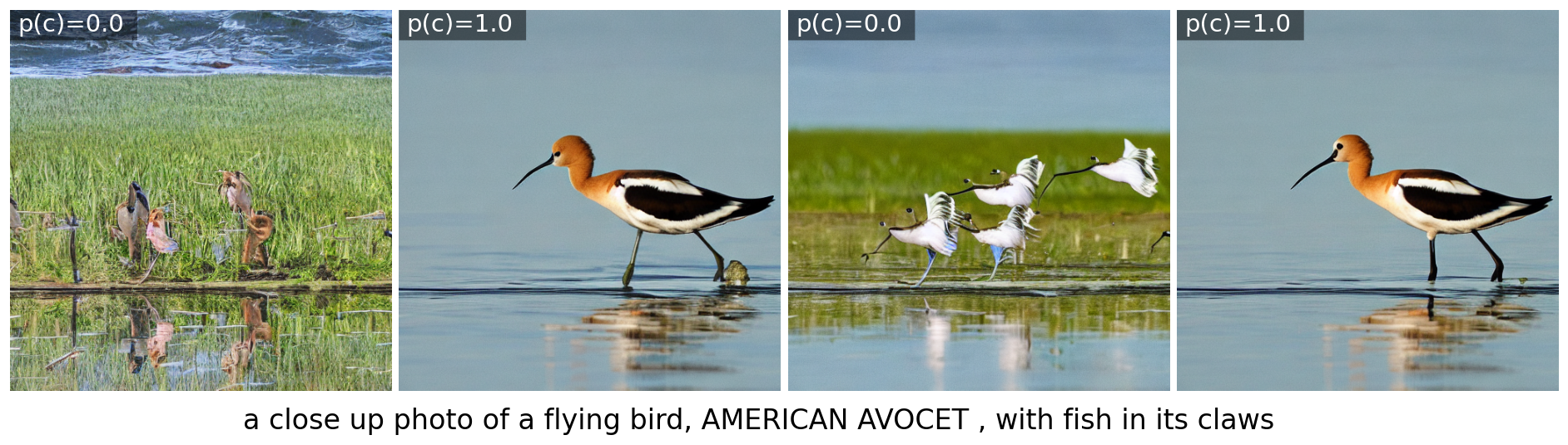}
      \caption{\hspace{1em} NG \hspace{10em} CFG \hspace{9em} GFCG  \hspace{8em} GFCG+CFG}
  \end{subfigure}
 \end{center}
 \vspace{-15pt}
 \caption{Representative visual examples where GFCG fails to improve class accuracy using detailed prompts.}
 \label{fig:dtpb}
\end{figure*}

\subsection{Detailed Text Prompts}
\begin{figure*}[t!]
\captionsetup[subfigure]{font=footnotesize, labelformat=empty}
 \begin{center}
  \begin{subfigure}[b]{0.74\textwidth}
    \centering
      \vspace{-0.1em}
      \includegraphics[width=\textwidth, interpolate=false]{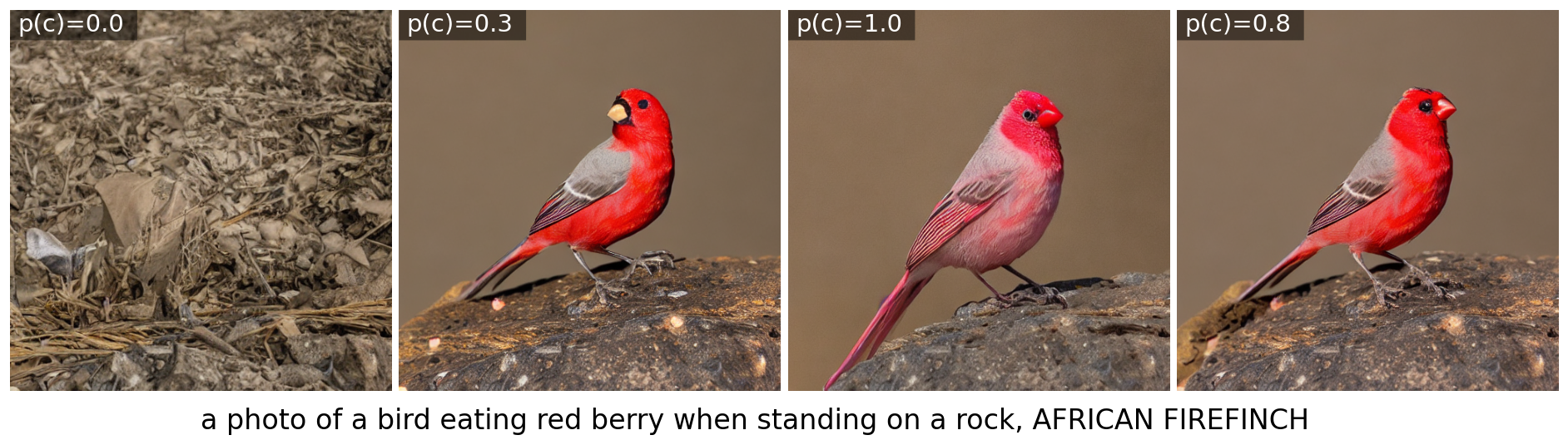}
  \end{subfigure}
  \begin{subfigure}[b]{0.74\textwidth}
    \centering
      \vspace{-0.1em}
      \includegraphics[width=\textwidth, interpolate=false]{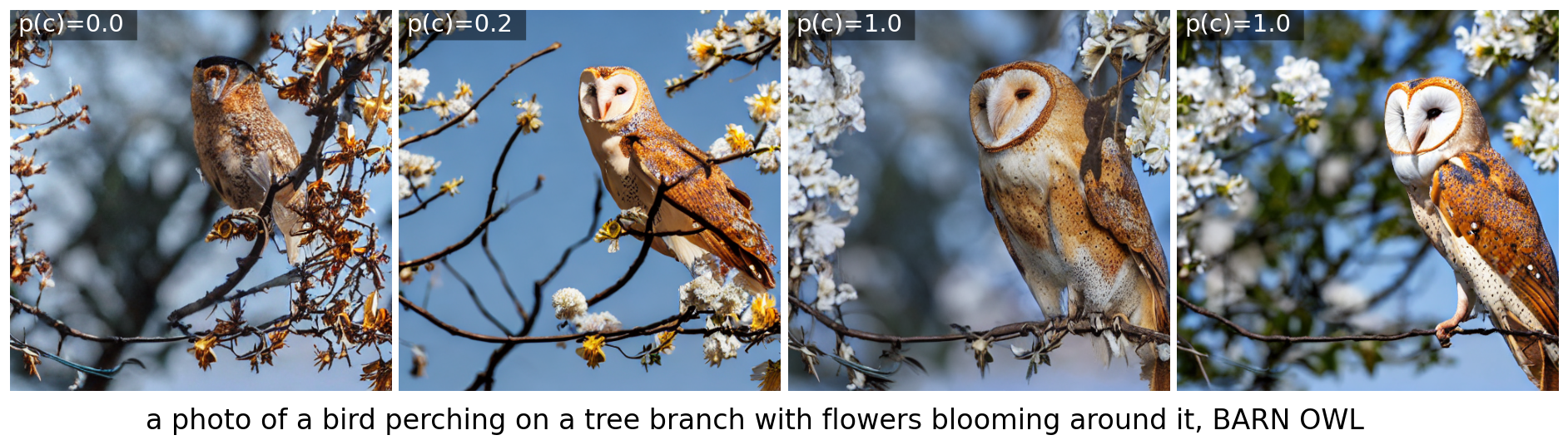}
  \end{subfigure}
  \begin{subfigure}[b]{0.74\textwidth}
    \centering
      \vspace{-0.1em}
      \includegraphics[width=\textwidth, interpolate=false]{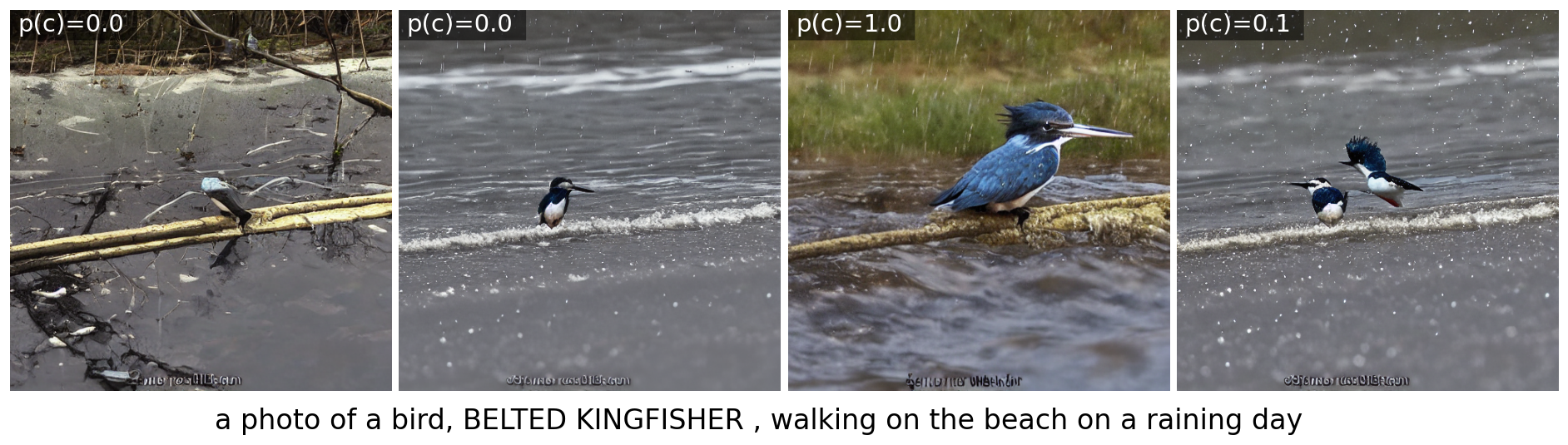}
  \end{subfigure}
  \begin{subfigure}[b]{0.74\textwidth}
    \centering
      \vspace{-0.1em}
      \includegraphics[width=\textwidth, interpolate=false]{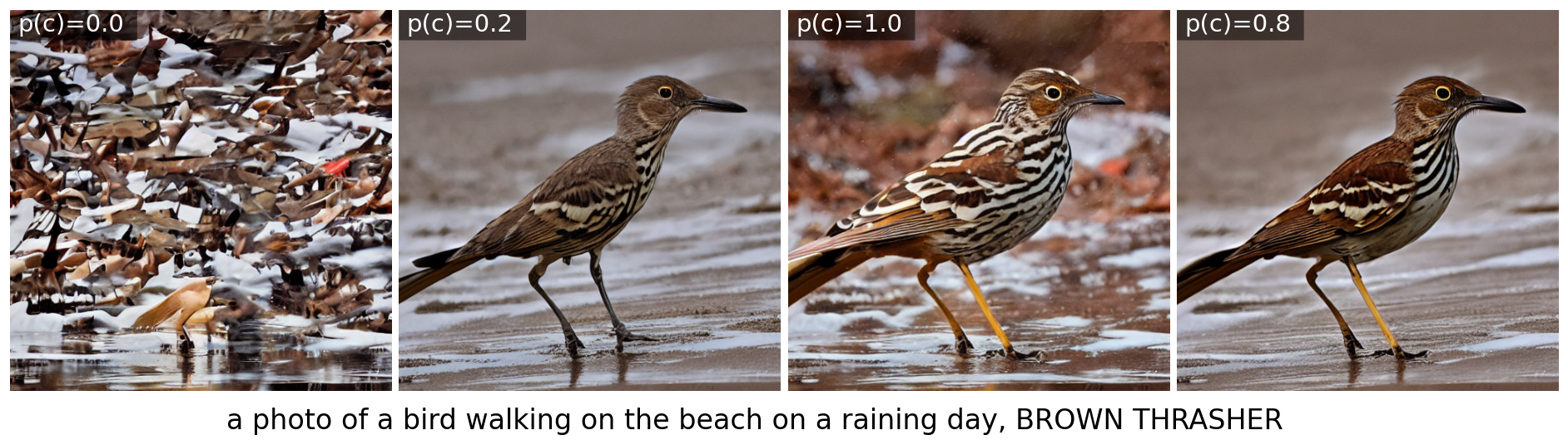}
  \end{subfigure}
  \begin{subfigure}[b]{0.74\textwidth}
    \centering
      \vspace{-0.1em}
      \includegraphics[width=\textwidth, interpolate=false]{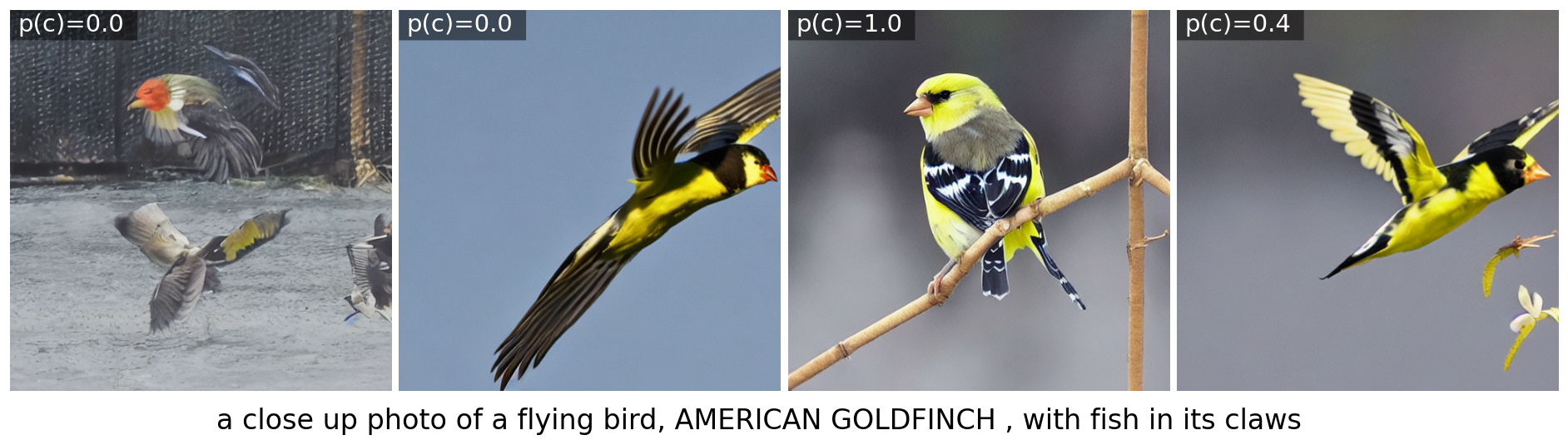}
  \end{subfigure}
  \begin{subfigure}[b]{0.74\textwidth}
    \centering
      \vspace{-0.1em}
      \includegraphics[width=\textwidth, interpolate=false]{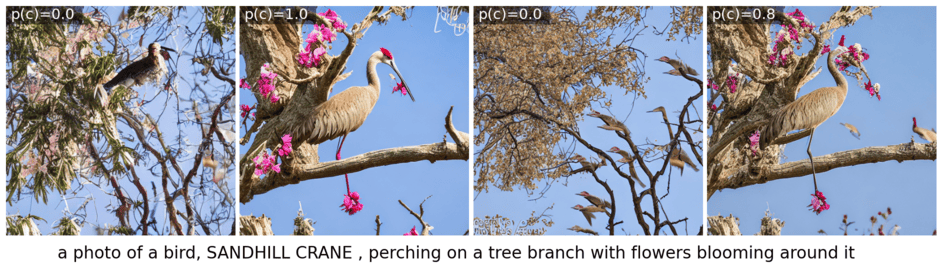}
      \caption{\hspace{1em} NG \hspace{10em} CFG \hspace{9em} GFCG  \hspace{8em} GFCG+CFG}
  \end{subfigure}
 \end{center}
 \vspace{-15pt}
 \caption{More visual examples from SD 1.5 model using detailed text prompts.}
 \label{fig:dtpm}
\end{figure*}

Some visual examples using the set of detailed prompts are included in Figure ~\ref{fig:dtpg}.  Note that the hyperparameters like $\alpha$ and $t_s$ for GFCG related methods were optimized for the generic prompts and adopted for detailed prompts without further tuning.  It shows that GFCG has the best class accuracy while preserving the overall accuracy of the full text prompt, including improving large features like the first example, or small details like the last one.  It is noted that, all method including CFG, have difficulty in depicting some details in the prompts like \textit{fish in its claws} and \textit{eating red berry}. As shown in Figure \ref{fig:dtpb}, for failure cases of GFCG where class probabilities of GFCG results are lower than those of CFG, certain features of the species, e.g. white feather of \textit{BALD EAGLE}, are excessively enhanced.  This could be caused by improper guidance scales. More visual examples using detailed text prompts are shown in Figure \ref{fig:dtpm} at the end.

\begin{figure*}[t!]
\captionsetup[subfigure]{font=footnotesize, labelformat=empty}
 \begin{center}
  \begin{subfigure}[b]{0.8\textwidth}
    \centering
      \vspace{-0.1em}
      \includegraphics[width=\textwidth, interpolate=false]{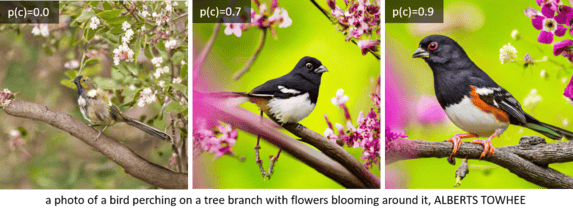}
  \end{subfigure}
\iffalse
  \begin{subfigure}[b]{0.8\textwidth}
    \centering
      \vspace{-0.1em}
      \includegraphics[width=\textwidth, interpolate=false]{CVPR2025/images/if_supp_samples/if_1.png}
  \end{subfigure}
  \begin{subfigure}[b]{0.8\textwidth}
    \centering
      \vspace{-0.1em}
      \includegraphics[width=\textwidth, interpolate=false]{CVPR2025/images/if_supp_samples/if_2.png}
      % \caption{\hspace{1em} NG \hspace{10em} CFG \hspace{9em} GFCG  \hspace{8em} GFCG+CFG}
  \end{subfigure}
  \begin{subfigure}[b]{0.8\textwidth}
    \centering
      \vspace{-0.1em}
      \includegraphics[width=\textwidth, interpolate=false]{CVPR2025/images/if_supp_samples/if_3.png}
      \caption{\hspace{3em} NG \hspace{14em} CFG \hspace{13em} GFCG+CFG}
  \end{subfigure}
\fi

%  \end{center}
%  \label{fig:if_samples_a}
% \end{figure*}\

% \begin{figure*}[t!]
% \captionsetup[subfigure]{font=footnotesize, labelformat=empty}
%  \begin{center}
  \begin{subfigure}[b]{0.8\textwidth}
    \centering
      \vspace{-0.1em}
      \includegraphics[width=\textwidth, interpolate=false]{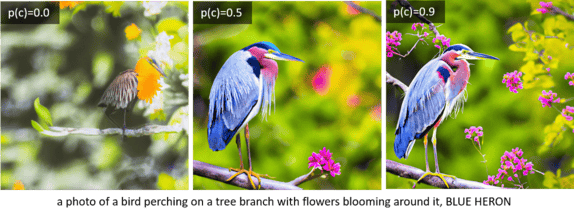}
  \end{subfigure}
  \begin{subfigure}[b]{0.8\textwidth}
    \centering
      \vspace{-0.1em}
      \includegraphics[width=\textwidth, interpolate=false]{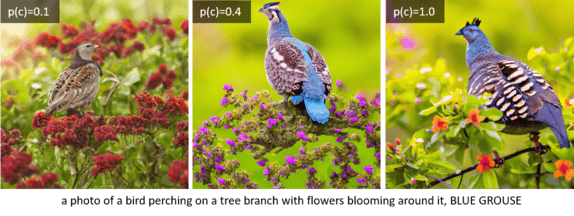}
  \end{subfigure}
  \begin{subfigure}[b]{0.8\textwidth}
    \centering
      \vspace{-0.1em}
      \includegraphics[width=\textwidth, interpolate=false]{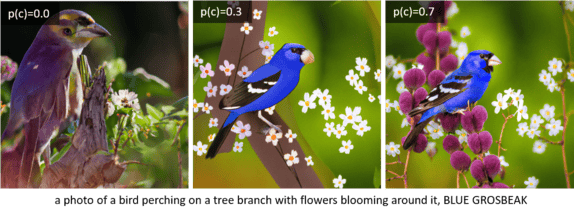}
      \caption{\hspace{3em} NG \hspace{14em} CFG \hspace{13em} GFCG+CFG}
  \end{subfigure}
 \end{center}
 \vspace{-15pt}
 \caption{Visual examples which demonstrate the benefits of GFCG in text-to-image generation using pixel-level diffusion model (i.e. without latent diffusion model). The test was performed on the DeepFloyd IF model from Stability AI. GFCG improves the class accuracy of generated image for multiple bird species, by aligning visual presence of the bird to the actual appearance observed in real world.}
 \label{fig:if_samples_b}
\end{figure*}

\subsection{Text-to-image Generation in Pixel Space}
\label{sec:if}
\begin{figure*}[t!]
\captionsetup[subfigure]{font=footnotesize, labelformat=empty}
 \begin{center}
  \begin{subfigure}[b]{0.8\textwidth}
    \centering
      \vspace{-0.1em}
      \includegraphics[width=\textwidth, interpolate=false]{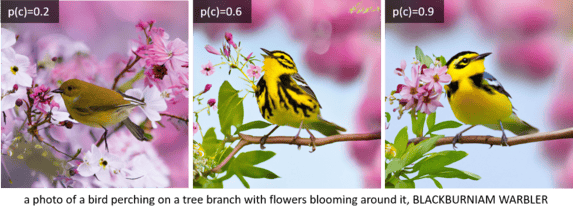}
  \end{subfigure}
  \begin{subfigure}[b]{0.8\textwidth}
    \centering
      \vspace{-0.1em}
      \includegraphics[width=\textwidth, interpolate=false]{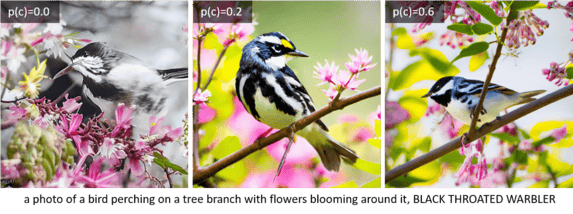}
      % \caption{\hspace{1em} NG \hspace{10em} CFG \hspace{9em} GFCG  \hspace{8em} GFCG+CFG}
  \end{subfigure}
  \begin{subfigure}[b]{0.8\textwidth}
    \centering
      \vspace{-0.1em}
      \includegraphics[width=\textwidth, interpolate=false]{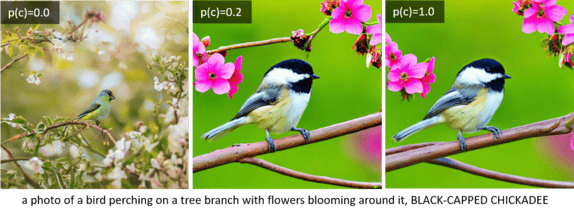}
      \caption{\hspace{3em} NG \hspace{14em} CFG \hspace{13em} GFCG+CFG}
  \end{subfigure}
 \end{center}
 \vspace{-15pt}
 \caption{More visual examples from DeepFloyd IF model using detailed text prompts.}
 \label{fig:if_samples_m}
\end{figure*}

We also explored using GFCG in the pixel space alone, without using latent diffusion like SD 1.5 used above. In this test, GFCG is integrated into the DeepFloyd IF model from Stability AI, whose diffusion mechanism is implemented in the pixel level. The IF model is able to generate high definition images in size of $1024\times1024$ based on given text prompts. We assessed the performance on the Bird Species dataset, and showed some visual examples in Figure ~\ref{fig:if_samples_b}.  The probability of classifying each generated sample as the target class is also included for reference.  On multiple bird species, much higher class accuracy was achieved by GFCG over other guidance schemes.  The gained accuracy is largely because of enhanced visual quality of the bird, which is more aligned to the actual appearance of corresponding species.  Some additional examples are shown in Figure \ref{fig:if_samples_m} at the end.

\end{document}